\documentclass{bmvc2k}

\usepackage{amsmath,amssymb}   % amssymb -> \mathbb
\usepackage{booktabs}
\usepackage{multirow}
\usepackage{wrapfig}
\usepackage{graphicx}
\usepackage{xcolor}
\usepackage{enumitem}
\usepackage{xspace}
\usepackage{pifont}
\usepackage{amsthm}
\usepackage{siunitx}
\usepackage{subfigure}   % supplementary uses \subfigure
\usepackage{diagbox}

\newtheorem{proposition}{Proposition}

\definecolor{m_green}{HTML}{2C8915}
\definecolor{m_red}{HTML}{FF0000}

\newcommand{\Tin}{T_{\text{in}}}
\newcommand{\Tout}{T_{\text{out}}}
\newcommand{\cten}{c_{10}}
\newcommand{\fr}{f_{r}}
\newcommand{\fg}{f_{g}}

\title{Real-time body pose non-verbal communication with a consistency-based reliability measure}

\addauthor{Alina Marcu}{alina.marcu@upb.ro}{1}
\addauthor{Drago\c{s} Costea}{dragos.costea@upb.ro}{1}
\addauthor{Cristina Laz\u{a}r}{cristina.lazar@upb.ro}{1}
\addauthor{Marius Leordeanu}{marius.leordeanu@upb.ro}{1,2,3}

\addinstitution{
National University of Science and Technology "Politehnica" Bucharest\\
}
\addinstitution{
"Simion Stoilow" Institute of Mathematics of the Romanian \\ Academy
}
\addinstitution{
NORCE Norwegian Research \\ Centre AS
}

\runninghead{Marcu, Costea, Lazar, Leordeanu}{Non-verbal Body Pose Communication}
\begin{document}

\maketitle

\begin{abstract}
\iffalse
Recognizing non-verbal communication from body pose is increasingly important for many use cases that require real time low cost person-to-robot communication in long distance environments. Body pose could be the only means of communication when other modalities beyond vision are not available. There is no dataset in the literature appropriate for this task, most are either multi-modal or require high resolution input. The body pose ones are limited to specific actions (running, walking, etc) and are not meant for person-to-machine communication. We introduce two datasets for pose-only communicative intent across ten balanced classes - a real world one captured with human actors and a generated one (Veo 3.1), and prove their value in extensive comparisons with other real datasets (Seamless Interaction) and existing synthetic generation methods (MotionLCM). We train a lightweight, embed-compatible dual-task model that, at each step, classifies the intent and predicts the next pose. From far away and limited data, the classification might be inconsistent. We introduce a novel unsupervised consistency based measure for assessing test time accuracy that feeds the prediction of the model back into itself. From this autoregressive output, we derive a label-free reliability plot, shown in experiments to discriminate in model robustness characterization.
\fi
Body movement communicates intent at distances and in conditions where neither the face, nor speech can be captured. We study the recognition of communicative intent from 2D body pose alone. We argue that body motion is a reliable signal especially in scenarios that require real time low-cost on-device person-to-robot communication in long distance environments, such as rescue missions. However, existing resources do not isolate this signal. Affective corpora combine body, face, voice and text, while skeleton action-recognition benchmarks label the action performed rather than the message conveyed. We release a dataset of real frames of full-body pose covering ten communicative intents and we compare it against other real (IPC) and synthetic (MotionLCM, VEO3.1, Kimodo) ones that span a range of difficulty. We target systems that can run on a robot's limited onboard hardware. We benchmark multiple models, from skeleton graph classifiers to joint motion-forecasting networks, and report performance metrics together with frame rate on an embedded GPU (NVIDIA Orin~Nano), since speed matters as much as accuracy in our scenario. Finally, we show that a model's own \emph{autoregressive self-consistency} works as an unsupervised reliability signal. We give a short proof that bounds the probability that a self-consistent prediction is correct, show that this probability grows with the number of consistent steps, and identify the conditions under which a confident prediction can still be false, benchmarked against industry-standard metrics.
\end{abstract}

%-------------------------------------------------------------------------
\section{Introduction}
\label{sec:intro}

A cornerstone of human connection is non-verbal communication. Body language,
posture, and gestures often convey emotional information (or intent) that words alone do not~\cite{rivera2024emotion,hazmoune2024transformers}. This silent channel becomes the only channel in a range of practical settings, such as a person signaling to a mobile rescue robot across an open field, an operator directing a ground robot in a noisy industrial bay, a camera that must respect privacy by not recording faces, or a rescue scenario in which speech is impossible and only a distant body is visible~\cite{perera2021uav}. In all of the above the face is too small to read, audio is unavailable, and context is absent, yet the body is still expressive. Our question is whether communicative \emph{intent} can be recovered from 2D body pose alone, on hardware small enough to sit on the robot that must react to it.

The field is not set up to answer this. Affective corpora such as MELD~\cite{poria2018meld} and IEMOCAP~\cite{busso2008iemocap} entangle body motion with speech, face, and dialogue, so the body-only signal cannot be isolated. Skeleton action-recognition benchmarks such as NTU~RGB+D~120~\cite{liu2020ntu120} label
\emph{what action} is performed (walking, sitting, waving) over multi-second
windows, not \emph{what is communicated} in the one-to-two-second bursts that
carry intent. Drone-gesture systems \cite{perera2021uav,kim2025drone} do use
body-only input but reduce it to a small fixed vocabulary of command poses, not
the richer and natural space of communicative intents. The result is a genuine gap: no
dataset or benchmark targets the recognition of communicative intent from body
pose alone, under the real-time, on-device constraints that challenging applications impose.

Recovering intent is only half of what a deployable system needs. A predictor
inside a control loop must also signal \emph{when it is likely to be wrong}, so
that a misread intent can be deferred rather than acted upon (the failure that matters most in a safety-relevant setting). This connects to the literature
on calibration and selective prediction: modern networks are poorly calibrated
and overconfident on their errors~\cite{guo2017calibration}, and methods such as SelectiveNet~\cite{geifman2017selective} improve reliability by abstaining when uncertain.
Most such methods, however, need labelled validation data to calibrate. We ask
instead whether a model can derive a reliability signal from its own behaviour at
test time, with no labels at all, by measuring how consistently it classifies its
own forecasted motion as it is fed back over an autoregressive rollout.

This paper makes three main contributions:
\vspace{-0.5em}
\begin{itemize}
\item We release a dataset for recognizing communicative intent from
body pose: $157k$ real frames over 10 intents. We evaluate it alongside four further corpora (one real, three synthetic) chosen to span a range of difficulty, so that differences
between models can be attributed to the models rather than to a single dataset (Section~\ref{sec:datasets}).
\vspace{-0.5em}
\item We benchmark twelve models under matched training conditions, covering
classification-only skeleton graph networks and joint forecasting/recognition
networks, and we report frame rate on an embedded GPU together with accuracy
(Sections~\ref{sec:method},~\ref{sec:experiments}). The contribution does not claim novelty for any single architecture, but a comparison and the
accuracy/latency picture it gives.
\vspace{-0.5em}
\item We study a label-free confidence estimate based on a model's agreement with
itself along an autoregressive rollout. We give a short proof bounding the
probability that a self-consistent prediction is correct, show that it grows with
the number of consistent steps, and identify when a confident prediction can
still be wrong. The predicted behaviour, including the failure mode, appears
across all tested datasets (Section~\ref{sec:experiments}).
\end{itemize}

%-------------------------------------------------------------------------
\section{Related Work}
\label{sec:related}

% DONE
We position our work at the gap between action recognition, which asks what a body \emph{does}, and communication understanding, which asks what a body
\emph{means}. Although recognizing actions from pose is mature, classifying communicative intent from body pose alone (in real-time, signal-denied
settings such as person-to-robot interaction) remains unaddressed. We further contribute an unsupervised, consistency-based reliability estimate, which we
relate to the calibration literature below.

% DONE
\noindent\textbf{Action recognition from pose.} Skeleton-based action recognition has matured significantly. Early datasets like MovingPose~\cite{zanfir2013moving} pioneered pose-based recognition, while NTU RGB+D 120~\cite{liu2020ntu120} now contains 120 action classes with models exceeding 90\% accuracy. On the modeling side, ST-GCN~\cite{yan2018stgcn} pioneered spatial-temporal graph convolutions over skeleton sequences, with subsequent advances including adaptive graph topologies (2s-AGCN~\cite{shi2019twostreamadaptive}), channel-wise topology refinement (CTR-GCN~\cite{chen2021ctrgcn}), and transformer-based methods such as ViTPose~\cite{xu2024vitpose} and MotionBERT~\cite{zhu2023motionbert}. However, these datasets and models focus on \textit{what} action is performed (walking, sitting, waving) rather than \textit{why} or \textit{how} it communicates intent. The temporal windows typically span 5--10 seconds for complete action cycles, whereas communicative gestures often occur in 1--2 second bursts. This distinction is crucial: recognizing ``waving'' (action) differs fundamentally from understanding whether the wave expresses greeting, farewell, or distress (communication). While action recognition excels at the former, no existing dataset addresses the latter when only body pose is available, which is a common constraint in drone surveillance, long-distance observation, and privacy-preserving applications.

% DONE
\noindent\textbf{From multimodal affect to pose-only intent.} Early studies established that non-verbal behaviors are critical for emotional dynamics~\cite{cristescu2008emotions}, with subsequent research showing these cues significantly impact user experience~\cite{cafaro2016first}. The survey by Noroozi et al.~\cite{noroozi2021emotional} comprehensively reviews emotional body gesture recognition, showing that while person detection and pose estimation are now mature technologies, emotion recognition from body gestures alone suffers from scarce labeled data and no consensus on output spaces. Existing systems invariably rely on rich multimodal inputs: Saunderson and Nejat~\cite{saunderson2019robots} use facial expressions alongside body language; Urakami et al.~\cite{urakami2023nonverbal} and Lozano-Hemmer~\cite{lozano2023open} assume access to contextual information beyond pose. The reliance on multimodal data limits deployment in real-world scenarios. Drones operating at 50+ meters cannot capture facial expressions; security cameras must respect privacy by avoiding face recording; communication with non-human primates cannot rely on speech. Our work deliberately constrains itself to skeletal keypoints alone, forcing the model to extract communicative intent purely from body dynamics, which is a significantly harder problem that reflects real-world constraints.

% DONE
\noindent\textbf{Body-gesture interfaces for mobile robots.} The use of body gestures as a natural interface for unmanned aerial vehicles has received growing attention. Perera et al.~\cite{perera2021uav} proposed an on-board UAV rescue system combining OpenPose skeleton extraction with a DNN classifier to recognize ten body rescue gestures from a drone's camera, demonstrating the viability of body-only communication in emergency scenarios where speech is impractical. The Skybound Magic framework~\cite{skybound2025} achieved body-only drone piloting at 30~FPS on a Jetson Nano, mapping eight canonical whole-body poses to flight commands. Kim et al.~\cite{kim2025drone} extended this paradigm using MediaPipe body-landmark detection for gesture-based IoT device control from a human-following drone. Cauchard et al.~\cite{cauchard2015drone} conducted foundational human-computer interface studies revealing which body gestures humans naturally choose for drone interaction, while Suarez-Fernandez et al.~\cite{suarez2016natural} explored natural user interfaces for human-drone multimodal interaction. A common limitation across these works is their reliance on small, hand-engineered gesture vocabularies with fixed command mappings, without addressing the richer problem of communicating emotions, intentions, or nuanced states through body pose alone for natural interaction, which is what our work addresses.

% DONE
\noindent\textbf{Datasets vs. the constrained reality.} Existing human interaction datasets reveal a critical bias toward multimodal richness. MELD~\cite{poria2018meld} and IEMOCAP~\cite{busso2008iemocap} provide emotional labels inseparable from verbal dialogue. CMU-MOSEI~\cite{zadeh2018mosei} offers 23,453 annotated segments but relies on facial expressions and speech. Body-focused motion datasets like AMASS~\cite{mahmoud2019amass} lack emotion labels, while emotion-annotated datasets like GEMEP~\cite{banziger2012gemep} include facial and vocal channels that contaminate body-only analysis. Even the comprehensive NTU RGB+D 120~\cite{liu2020ntu120} focuses on daily activities rather than communicative intent. This multimodal bias has created a blind spot: no dataset exists for classifying communication types from short body pose sequences alone. Our datasets fill this gap with 10 communication categories expressed purely through body language, each clip constrained to $\sim$2 seconds to perceive communicative intent while maintaining real-time responsiveness.

\iffalse
\paragraph{Calibration and selective prediction.}
Our reliability estimate relates to uncertainty quantification. Guo et al.\
\cite{guo2017calibration} showed modern networks are miscalibrated and overconfident on
errors; selective-prediction methods such as SelectiveNet \cite{geifman2017selective}
abstain under uncertainty to raise accuracy on retained predictions. These
typically require a labelled validation set to fit the confidence rule. Our
$c_{10}$ score needs no labels and no auxiliary head: it reuses the model's own
forecasting rollout as the source of a confidence signal, and we characterise
analytically when that signal is trustworthy and when it is not.
\fi

%-------------------------------------------------------------------------
\section{Methodology}
\label{sec:method}

%This section summarizes the technical pipeline used in our study. It includes formal task definition, the neural architectures adopted for full-body motion forecasting and intent classification, and details about the datasets.

\subsection{Problem statement}
\label{sec:method:models}

Our objective is twofold: to forecast short-term body motion keypoints in real time and to infer the underlying affective state from observed body-keypoint trajectories so that downstream systems can adapt to fine-grained non-verbal motion patterns. Each model maps an observed window
$\mathbf{X}\in\mathbb{R}^{\Tin\times V\times C}$ ($V{=}17$ joints, $C{=}2$
coordinates) to a future trajectory and a class distribution,
\begin{equation}
(\hat{\mathbf{Y}},\,\mathbf{z})=f_\theta(\mathbf{X}),\qquad
\hat{\mathbf{Y}}\in\mathbb{R}^{\Tout\times V\times C},\quad
\mathbf{z}\in\mathbb{R}^{K},
\end{equation}
with predicted intent $\hat{y}=\arg\max_k z_k$ and posterior
$\hat{p}_k=\mathrm{softmax}(\mathbf{z})_k$. Dual-task models produce both outputs, whilst
classification-only models produce only $\mathbf{z}$, and the trajectory term is
dropped. Training minimizes a weighted sum of an $\ell_2$ loss on the future poses
and the cross-entropy on the label,
\begin{equation}
\mathcal{L}(\theta)=\lambda_{\text{kp}}\,
\frac{1}{|\mathcal{B}|}\sum_{i\in\mathcal{B}}
\big\lVert\hat{\mathbf{Y}}_i-\mathbf{Y}_i\big\rVert_2^2
+\lambda_{\text{cls}}\,
\frac{1}{|\mathcal{B}|}\sum_{i\in\mathcal{B}}\big(-\log\hat{p}_{i,y_i}\big),
\end{equation}
with $(\lambda_{\text{kp}},\lambda_{\text{cls}})=(1.0,0.3)$ for dual-task models
and $(0,1)$ for classification-only models.

\noindent\textbf{Implementation details.} Models were optimized under the same setup, so that differences in the results follow from the architecture rather than the training procedure. We used Adam optimizer to train the models for 200 epochs using mini-batches of 32 samples. Implementation relied on the PyTorch framework~\cite{paszke2017automatic}. All runs were carried out on a workstation running Ubuntu 22.04.5 LTS, fitted with an Intel Core i9-14900K processor, 188 GB of system memory, and a single NVIDIA RTX 4090 graphics card. However, the models were selected to be compact enough to operate above real-time rates on edge hardware (see Table~\ref{tab:speed_orin}).

\subsection{Self-consistency as a confidence estimate}
\label{sec:method:reliability}

A predictor used inside a control loop should also signal when it is likely to be
wrong. The forecasting head lets us obtain such a signal without extra
components. Starting from an observed window, we append the model's own forecast
to the history and predict again, repeating to produce a long self-generated
sequence. At each step $\ell$ we re-classify the current window and record the
running accuracy
$\text{Acc}_\ell=\tfrac1\ell\sum_{m\le\ell}\mathbb{1}[\hat y^{(m)}{=}y]$. A model
that has learned the motion keeps generating in-class movement and loses accuracy
slowly. Less robust models tend to quickly drift to a poor classification accuracy.

From the first 10 predictions we define a label-free consistency score.

With
$\hat y^\star=\operatorname{mode}(\hat y^{(1)},\dots,\hat y^{(10)})$,
\begin{equation}
\cten=\frac{1}{10}\sum_{m=1}^{10}\mathbb{1}\!\left[\hat y^{(m)}=\hat
y^\star\right]\in[0.1,1.0].
\end{equation}
This needs no ground truth and is available at test time. The question is whether
it is calibrated: are the clips on which the model keeps predicting the same
label the clips on which that label is more often correct?

We can answer this with a short argument. Reduce the problem to two classes, the predicted label
and the single most-confused alternative, and let $n$ be the number of
consistent steps. Let $\fr>1$ be the factor by which the classifier favours the
correct label over the alternative on in-class motion, and $\fg>1$ the factor by
which the generator produces in-class over off-class motion. Under three
assumptions (A1) confusions are dominated by one alternative class, (A2) the
rollout steps are conditionally independent given the underlying hypothesis with
stationary per-step behaviour, and (A3) $\fr,\fg>1$, we obtain the following:

\begin{proposition}
\label{prop:main}
Under \textnormal{(A1)--(A3)}, the probability that an $n$-consistent prediction
is correct is
\begin{equation}
P(\textnormal{correct}\mid n)=\frac{1}{\,1+\fr^{-n}+(\fr\fg)^{-1}\,}.
\label{eq:prop}
\end{equation}
It lies in $(0,1)$ for all $n\ge1$ and $\fr,\fg>1$, increases strictly with $n$,
and tends to
\begin{equation}
P_\infty=\lim_{n\to\infty}P(\textnormal{correct}\mid n)=\frac{\fr\fg}{1+\fr\fg}<1.
\label{eq:limit}
\end{equation}
\end{proposition}

Equation~\eqref{eq:prop} is bounded in $(0,1)$ by construction, since it has the
form $1/(1+\text{positive})$; this corrects an earlier version of ours in which
the bound was not guaranteed. The probability rises with $n$, in line with the
intuition that more agreement should mean more reliability. The informative part
is the limit~\eqref{eq:limit}: consistency cannot raise the probability to $1$,
only to $\fr\fg/(1+\fr\fg)$. When the alternative class is easily confused
($\fr,\fg\to1^{+}$) this limit approaches $\tfrac12$, so a fully consistent
prediction is no better than a guess. This is the regime in which a model is
confident and wrong, which a confidence estimate has to be honest about. In
short, $\cten$ is informative when $\fr\fg$ is large -- the classes are well
separated in both recognition and generation -- and uninformative when
$\fr\fg\approx1$. Section~\ref{sec:experiments} shows both cases, namely the four
learnable datasets follow~\eqref{eq:prop}, and IPC, where recognition is near
chance and so $\fr\approx1$, follows~\eqref{eq:limit}.

\subsection{Datasets}
\label{sec:datasets}

We use a mix of real and synthetic datasets. We focus on the text-to-keypoints or text-to-video models. We focus on 10 intents: \textit{affection, come to me, dancing, disapproval, enthusiastic, go away, happy to see you, laughing, surprise, we have a deal}. They were chosen as a general vocabulary for someone that would like to communicate with an UAV/UGV for practical purposes, without specialized domain knowledge (i.e., NATOPS aircraft marshaling signs).

Although synthetic datasets can generate an (unlimited) number of samples, they lack the variability of the real ones and are generally limited to a single understanding of an intent. For example, \textit{come to me} is understood by VEO 3.1~\cite{veo2024} as a beckoning finger, which limits the usability compared to a real live scenario where a more ample body gesture as seen in the first figure (ours).

We evaluate body-language classification on five datasets that share a common interface: every sample is a 105-frame sequence of $17$ 2-D body keypoints in the COCO~\cite{lin2014microsoft} convention, sampled at 30Hz, stored as a $(N,\,105,\,17,\,2)$ tensor of pixel coordinates with a per-sequence categorical label. Except MotionLCM, all models sample 60 frames from this sequence with a sliding window. The datasets differ in (i) how the motion is sourced (real video vs.\ synthetic), and (ii) how the keypoints are recovered (off-the-shelf detector vs.\ direct output of a parametric body model). Their sizes and label distributions are summarized in Table~\ref{tab:datasets} and a detailed description follows.

\begin{wrapfigure}{R}{0.6\textwidth}
    \vspace{-21mm}

    \centering
    \includegraphics[width=0.58\textwidth]{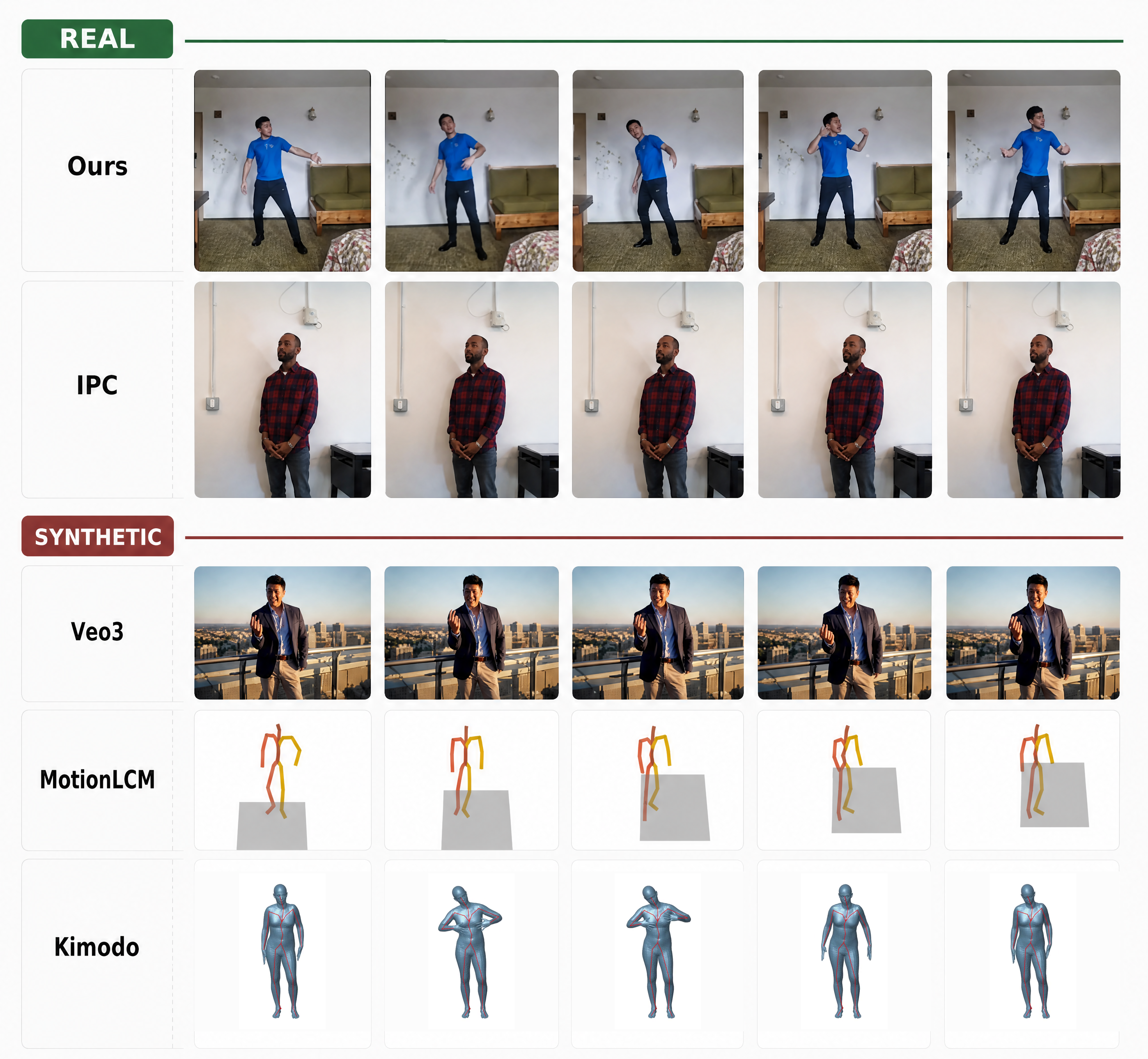}
    \vspace{-1em}
    \caption{Visual comparison of the \textit{come to me} intent on 4 datasets and a similar emotion from the IPC dataset - APCP (Assertive and Affiliation Positive). Ours exhibits
  large body motions compared to synthetic alternatives, while the sample from IPC is more focused on the face for the intent, rather than ample movement.}
    \label{fig:datasets_come_to_me}
    \vspace{-4mm}

  \end{wrapfigure}

\begin{table}[t]
\centering
\small
\caption{Datasets used in this work. Final number of frames of
$17$~COCO keypoints in $2$D. ``Train'' / ``Test'' give the
total number of \emph{frames} in each split;
``Classes' is the number of distinct labels.}
    \vspace{3mm}

\setlength{\tabcolsep}{4pt}
\begin{tabular}{llrrr}
\toprule
\textbf{Dataset} & \textbf{Source} & \textbf{Train} & \textbf{Test} & \textbf{Classes} \\
\midrule
Ours      & Real human video (YOLO11-pose~\cite{jocher2024yolo11})            & $125{,}475$    & $31{,}395$    & $10$ \\
IPC       & Real human video (YOLO11-pose)                                    & $100{,}800$    & $25{,}200$    & $8$  \\
MotionLCM & Latent consistency motion model~\cite{dai2024motionlcm}            & $8{,}400{,}000$ & $2{,}100{,}000$ & $10$ \\
Kimodo    & SMPL-X (Kimodo-SMPLX-RP-v1~\cite{rempe2026kimodo})  & $126{,}000$    & $31{,}500$    & $10$ \\
VEO3.1   & Text-to-video (Veo~3.1~\cite{deepmind2025veo}) + YOLO11-pose      & $100{,}800$    & $25{,}200$    & $10$ \\
\bottomrule
\end{tabular}

\label{tab:datasets}
\end{table}

% old figure

\iffalse
\begin{figure}[htbp]
\centering
\includegraphics[width=\textwidth]{images/bmvc/datasets_main.png}
\caption{Visual comparison of the \textit{come to me} intent on 4 datasets and a similar emotion from the IPC dataset - APCP (Assertive and Affiliation Positive). Ours exhibits large body motions compared to synthetic alternatives, while the sample from IPC is more focused on the face for the intent, rather than ample body movement.}
\label{fig:datasets_come_to_me}
\end{figure}
\fi

\begin{enumerate}

    \item \textbf{Ours (real)}
    Our dataset features a total of 156,870 frames, split across a set of 10 intents (affection, come to me, dancing, disapproval, enthusiastic, go away, happy to see you, laughing, surprise, we have a deal), balanced across each emotion and captured from 3 different participants. We recorded 1920x1080px 30fps videos in an indoor environment with a small number of participants, self-directed from the prompt. As opposed to existing datasets, ours features large body movements that can be easily recognized without the need for facial expressions or additional communication channels (i.e., voice).

\item \textbf{Seamless interaction -- IPC (real)}
    The Seamless Interaction dataset~\cite{agrawal2025seamless}, released by Meta, is a large-scale multimodal corpus comprising over 4,000 hours of face-to-face interaction footage recorded from more than 4,000 participants across diverse conversational contexts. The dataset spans two primary conditions: improvised interactions with professional actors guided by scripted prompts, and naturalistic spontaneous conversations, providing broad coverage of human communicative behavior. Each recording is annotated with body keypoints, voice activity detection labels, and human annotations of internal states and behavioral assessments. The 27 terabytes dataset is designed to support research in embodied AI and virtual agent synthesis. Although this is not a direct match to ours, it is the largest dataset that focuses on communicating an intent and it has 8 classes that follow the Interpersonal Circumplex (IPC). We refer this dataset as IPC in the following materials, due to its labeling.

\item \textbf{MotionLCM (synthetic)}
MotionLCM is a real-time human motion generation framework that extends latent diffusion models with latent consistency model (LCM) techniques to enable one-step or few-step
  inference, substantially reducing the computational overhead traditionally associated with diffusion-based motion synthesis \cite{dai2024motionlcm}. Its efficiency gains make it particularly well-suited for interactive applications such as real-time character animation and virtual agent control, where diffusion
  models were previously too slow to be practical. We use this model to generate 10M frames, split into the same 10 emotions, prompted as \textit{a person <intent>}, in a bid to determine whether low accuracy synthetic data can help real world detection.

\item \textbf{Kimodo (synthetic)}
Kimodo~\cite{rempe2026kimodo}, released by NVIDIA, is a text-conditioned SMPL-X~\cite{pavlakos2019smplx} motion-generation model. We prompt it with ``\textit{a person showing X body language}'' for each of the ten emotions and sample $120$ motions per class with random seed $0$ to build the training set ($N{=}1{,}200$), then sample a further $30$~motions per class with seed $42$ to build the test set ($N{=}300$). Each motion has $105$ frames at $30$~Hz and is generated by Kimodo as a per-frame SMPL-X parameter sequence (global orientation, $21$ body-pose joints, root translation). We run the SMPL-X neutral forward pass to obtain $22$ posed joints in 3D and we project them to the $17$ COCO keypoints, matching the schema used by the other datasets.

\item \textbf{VEO3.1 (synthetic)} We generate 600 videos at the standard 1280x720px resolution using VEO3.1 Fast~\cite{veo2024}, totaling 115,200 frames. The same set of classes apply, the prompt being \textit{A person showing "<intent>" body language gesture, full body shot.} The prompt was slightly altered from MotionLCM due to empirical observations. Several classes had prompt adherence issues, i.e., "we have a deal" generally showed two persons shaking hands, but most results were similar in appearance with our dataset. Compared to Kimodo, these samples carry photorealistic appearance variation (lighting, clothing, scene context), making this dataset a useful intermediate between fully-synthetic skeletons (Kimodo, MotionLCM) and recorded human video (Ours, IPC).

\end{enumerate}

%-------------------------------------------------------------------------
\section{Experiments}
\label{sec:experiments}

We benchmark 12 models under the conditions presented in Section~\ref{sec:method:models}, spanning two categories. The first one, \emph{recognition-only}, includes 4 established skeleton action-recognition
architectures (ST-GCN++~\cite{duan2022pyskl},
CTR-GCN~\cite{chen2021ctrgcn}, 2s-AGCN~\cite{shi2019twostream}, EfficientGCN-B0~\cite{song2022efficientgcn}) and serves to
situate our task within the established skeleton-recognition literature. The
second category, \emph{joint forecast-recognition}, comprises of 4
baseline architectures without skeleton-specific structure (MLP, LSTM, CNN-LSTM, Transformer) together with four motion forecasters adapted for dual-task motion prediction and also classification
(MotionMixer~\cite{bouazizi2022motionmixer}, POTR~\cite{martinez2021potr},
InfoGCN++~\cite{chi2025infogcnpp}, PGBIG~\cite{ma2022pgbig}). The graph and forecaster baselines are reimplementations in reduced configurations, with smaller widths and depths than in their original papers. For full transparency we also present the details of the baseline models below:
\begin{itemize}
\item\textbf{MLP:} Multi-Layer Perceptron operating on the flattened input sequence. The 60 observed frames are reshaped into a vector of dimension $60 \times 17 \times 2 = 2040$. This vector is processed by a shared stack of four fully connected layers with 1024, 512, 256, and 128 units, respectively. Each layer is followed by ReLU, BatchNorm, and Dropout with $p=0.3$. The resulting 128-dimensional representation is passed to two output heads: a linear layer that predicts the flattened 30-frame future sequence, and a two-layer classifier for emotion recognition.
\item\textbf{LSTM:} This recurrent baseline consists of a two-layer LSTM with hidden size 128 and dropout 0.2. At each time step, the 17 keypoints are flattened into a 34-dimensional pose vector, yielding an input sequence of length 60. The last hidden state serves as the sequence representation and is forwarded to the generation and classification heads.
\item \textbf{CNN-LSTM:} This hybrid model first extracts local spatio-temporal patterns with a 1D convolutional frontend and then models longer dependencies with an LSTM. The input is transposed to shape [batch, 34, 60] and processed by three 1D convolutional layers with 64, 128, and 256 output channels and kernel size 3. Each convolution is followed by ReLU, BatchNorm, and Dropout. The resulting feature sequence is then encoded by a two-layer LSTM with hidden size 128. Its final hidden state is used by the two prediction heads.
\item \textbf{Transformer:} The Transformer variant uses an encoder-only design.
Each 34-dim frame descriptor is first projected into a latent space of dimension $d_{\text{model}}=256$. Sinusoidal positional encodings are added to preserve temporal order, after which the sequence is processed by a four-layer Transformer Encoder with eight attention heads. The representation associated with the final time step is used as the shared feature vector for motion forecasting and classification.
\end{itemize}

The five datasets are chosen to span a controlled difficulty axis rather than to
maximize any single number. \emph{Ours} (real, 10 intents) is the target
distribution. \emph{IPC} (Seamless Interaction, real, 8 interpersonal-circumplex
codes) is included as a deliberate negative control: a real interaction corpus
whose labels are \emph{not} expected to be recoverable from gross 2D body
motion. \emph{MotionLCM} and \emph{Kimodo} (synthetic) provide separability
ceilings under low- and balanced-variance generation respectively, and
\emph{VEO3.1} (synthetic, photorealistic) provides a high-variance synthetic
distribution.

\begin{figure}[htbp]
\centering
\includegraphics[width=\textwidth]{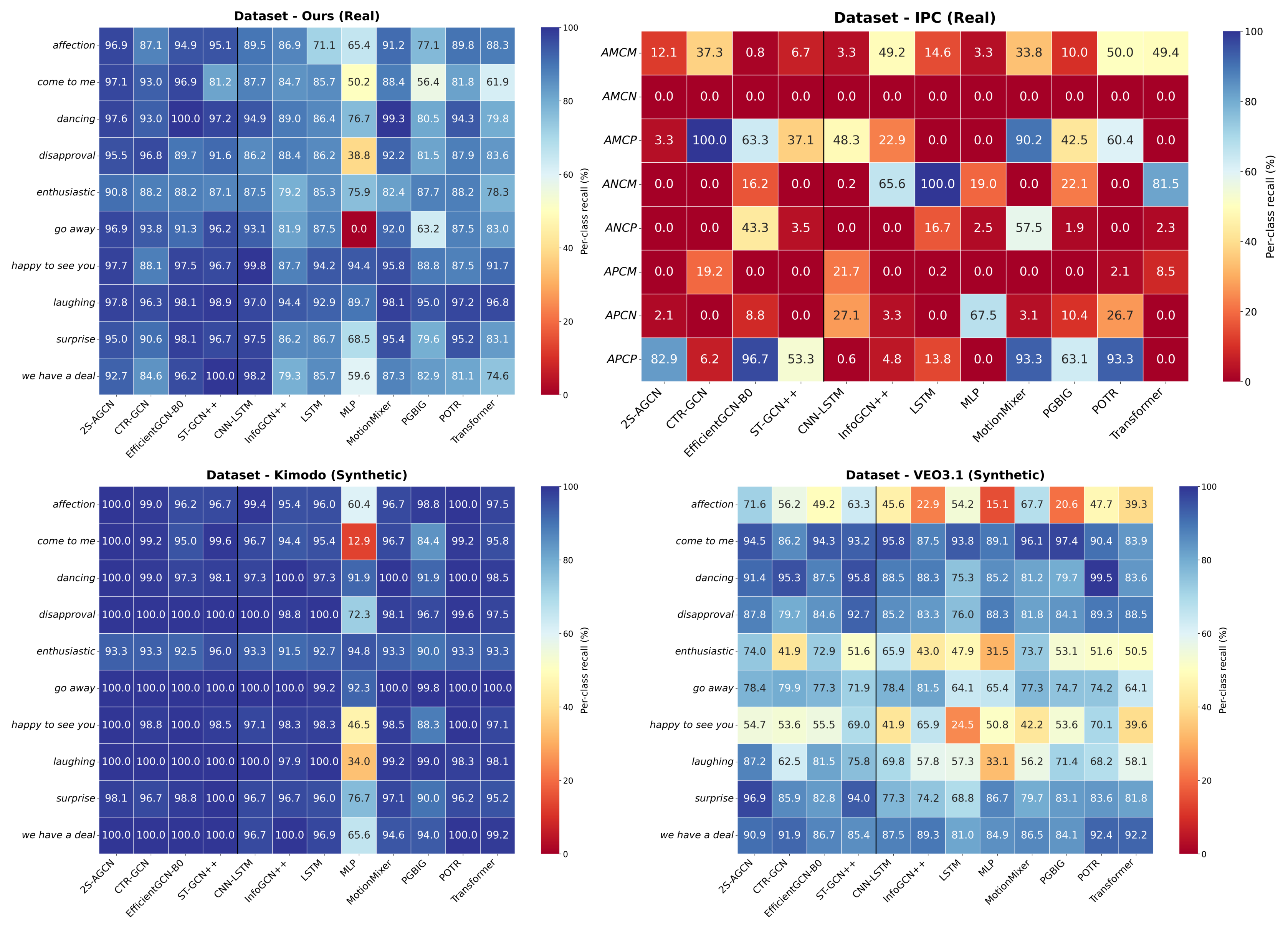}
\vspace{-2em}
\caption{Per-class recall (\%) on our proposed dataset, IPC, Kimodo and VEO3.1 (due to space limitations we moved the MotionLCM confusion matrix in the supplementary). Rows are intents (IPC carries its eight attribute codes). The methods run left to right as the four classification-only graph recognizers (2s-AGCN, CTR-GCN, EfficientGCN-B0, ST-GCN++) followed by the eight dual-task models. Blue is high recall, red is low.}
\label{fig:class_perf}

\end{figure}

\noindent\textbf{Per-class recognition across datasets.} Our evaluation is not asking \emph{how accurate} each model is on average, but
\emph{which intents each model fails to detect}. For an anticipatory
communication system, a missed intent is operationally worse than a hesitant
one, so per-class recall (measured as the fraction of test windows of true class $k$ that
the model assigns to $k$, $\mathrm{Recall}_k = M_{kk}/\sum_j M_{kj}$) is the
natural per-class analogue of aggregate accuracy (see Figure~\ref{fig:class_perf}). Within each heatmap, recognition-only and joint models are grouped and separated by a vertical rule. Results show that recognition is strong and largely uniform on the target distribution. On \emph{Ours}, the graph and recurrent models occupy the 90--100\% recall band on nearly every intent: 2S-AGCN, EfficientGCN-B0, ST-GCN++ and CNN-LSTM yield high recall, and the joint models track them closely. This establishes that body-only communicative intent is recoverable from 2D keypoints for the majority of our label space. The MLP is the only model whose column contains orange and red cells on \emph{Ours}, collapsing to 0.0\% recall on \emph{go away} and degrading sharply on \emph{disapproval} (38.8\%), \emph{come to me} (50.2\%) and \emph{we have a deal} (59.6\%). A small set of intents (\emph{come to me}, \emph{disapproval}, \emph{enthusiastic}) is hard for every model, which indicates genuine ambiguity of these gestures under 2D projection, rather than to any single modeling defect. The IPC panel is dominated by failed recognitions, most cells are at or near 0.0\% recall, and entire intents (e.g.\ \emph{AMCN}) are zero across nearly all models. On MotionLCM (supplementary) and Kimodo, the matrices are almost entirely in the
95--100\% band, with the \emph{sole} structured exception again being the MLP (e.g.\ Kimodo \emph{laughing} 34.0\%, \emph{come to me} 12.9\%; MotionLCM \emph{we have a deal} 38.4\%, \emph{go away} 48.7\%). Unlike the other synthetic sets, VEO3.1's photorealistic appearance and noisier trajectories spread the methods apart, making it the most informative recognition panel for comparing architectures. Two patterns stand out. The two-stream joint+bone prior of 2S-AGCN is the most robust under this noise, holding 78--97\% on most classes where other models scatter. And the hard classes are themselves informative: \emph{affection}, \emph{enthusiastic} and \emph{happy to see you} are pale-to-red for most models (affection falls to
15.1\% for MLP, 22.9\% for InfoGCN++, 20.6\% for PGBIG), identifying these low-energy, expression-driven states as the ones whose communicative content
least survives reduction to 2D body keypoints. MotionLCM and VEO3.1 share a synthetic origin yet produce opposite difficulty profiles is itself a result. Synthetic does not imply easy, and appearance realism translates into trajectory
variability that exercises architecture quality in a way low-variance generators
cannot.

\begin{figure}[htbp]
\centering
\includegraphics[width=\textwidth]{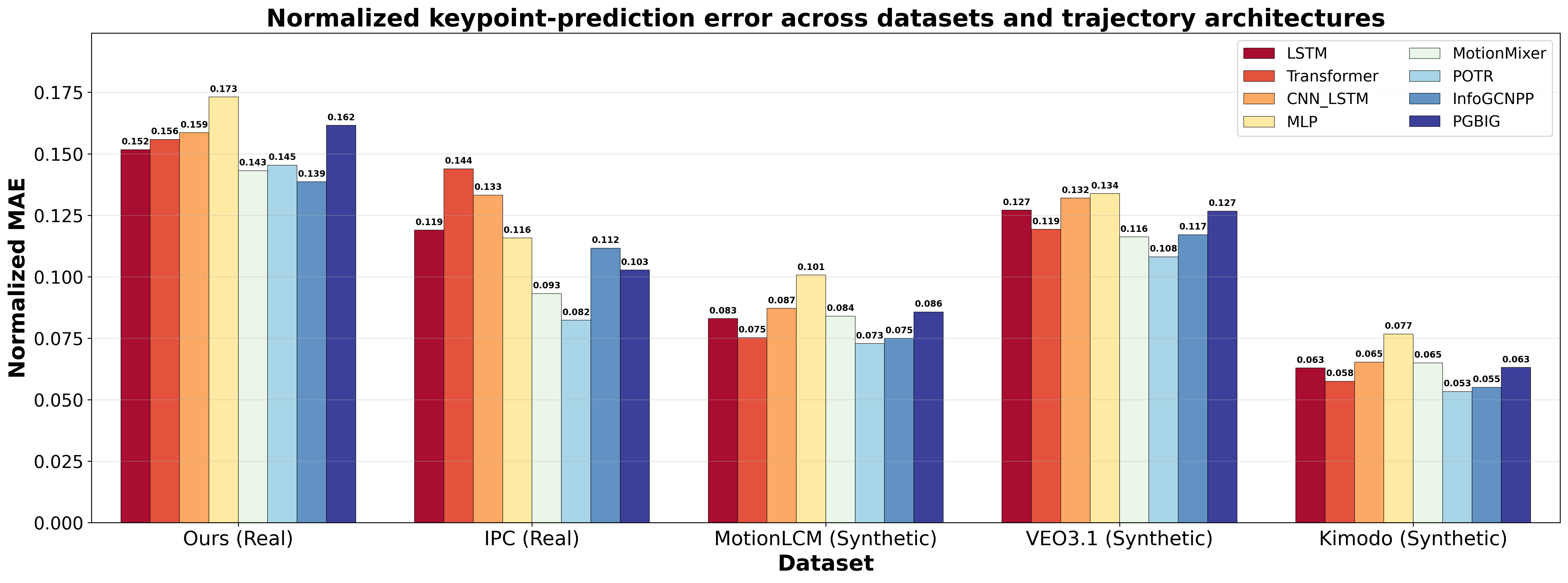}
\vspace{-2em}
\caption{Normalized keypoint-prediction error across the five datasets and the eight trajectory models (lower is better; true values printed above each bar).}
\label{fig:pred_perf}
\vspace{-1em}
\end{figure}

\noindent\textbf{Trajectory forecasting error.} For each test sequence, from each dataset, we compute a mean absolute error (MAE) over all 30 future frames, 17 joints and 2 coordinates, and normalize it by a per-sequence body-scale proxy $h_i$ (the mean absolute vertical coordinate of the ground-truth future), so that errors are comparable across datasets with different coordinate normalizations and can be placed on a single axis (see Figure~\ref{fig:pred_perf}). On the two real datasets the purpose-built motion models separate from the generic baselines. POTR and InfoGCN++ reach the lowest errors on IPC (0.082 and 0.139 respectively, against 0.119 for LSTM and 0.144 for the MLP and Transformer), and on \emph{Ours} the strongest forecasters (MotionMixer and POTR at 0.143, InfoGCN++ at 0.139) improve on the MLP's 0.173 by 17--20\%
relative. The ordering is the expected one. A non-autoregressive pose transformer with learned output queries (POTR), a future-prediction recurrent
model (InfoGCN++) and a temporal-mixing MLP designed for motion (MotionMixer) outperform sequence encoders that were never specialized for trajectory regression. The MLP records the highest error on four of the five datasets, exactly mirroring its recall collapse in Figure~\ref{fig:class_perf}. Without any reference to the recognition results, the bar heights recover the same difficulty ordering: the clean synthetic sets are easiest to predict (Kimodo 0.053--0.077, MotionLCM 0.073--0.101), the real \emph{Ours} is the
hardest (0.139--0.173), and VEO3.1 and IPC fall in between. However, low position MAE does not certify good motion. The metric rewards predictions that sit near the temporal mean of the future
trajectory. A model that emits an over-smoothed, nearly static rollout can therefore post a low MAE while producing motion that is physically implausible (frozen or regressed to the mean). The
over-smoothing failure mode motivates the autoregressive analyses which we regard as the substantive test of forecasting quality.

\noindent\textbf{Dataset diversity discussion}. We quantify diversity along three complementary axes. First, averaging per-class recognition recall across all twelve recognizers and all classes orders the datasets from most to least classifiable as Kimodo (94.84\%, with MotionLCM similar), Ours (87.15\%), VEO3.1 (72.67\%), and IPC (19.58\%). Second, the normalized keypoint-prediction MAE averaged over all eight trajectory architectures (lower is better) orders them Kimodo (0.062), MotionLCM (0.083), IPC (0.113), VEO3.1 (0.123), and Ours (0.154). Third, a within-distribution PCA experiment measures the average L2 distance between each test split and its training split (the residual variance after projecting the test set onto the components capturing 95\% of the training set's variance) ordering them MotionLCM (0.44), IPC (0.53), Ours (0.82), and VEO3.1 (0.93). Together these expose how varied the motion is within each dataset. The synthetic sets, Kimodo and MotionLCM, are easiest to recognize and predict because their motion is regular and repetitive in both splits. Their small PCA distances confirm that test samples are essentially template variations of the training ones. Real human motion is not like that. Ours is recognized almost as well but is the hardest to predict, as expected from movement too varied to extrapolate from 2D pose alone, and its larger PCA distance reflects a basic fact: no one performs the same emotion twice in the same manner. VEO3.1 sits in between, closer to reality than its synthetic counterparts. IPC is the outlier, focused on close-range, multimodal cues centered on the upper body and face rather than full-body motion. Without meaningful dynamic movement its skeleton trajectory carries little emotional signal over distance, leaving a body-kinematics-only model almost nothing to work with. Overall, low classification and prediction error track how simple and repetitive a dataset's motion is, while the PCA distances expose how much genuine diversity it contains.

\begin{figure}[htbp]
\centering
\includegraphics[width=\textwidth]{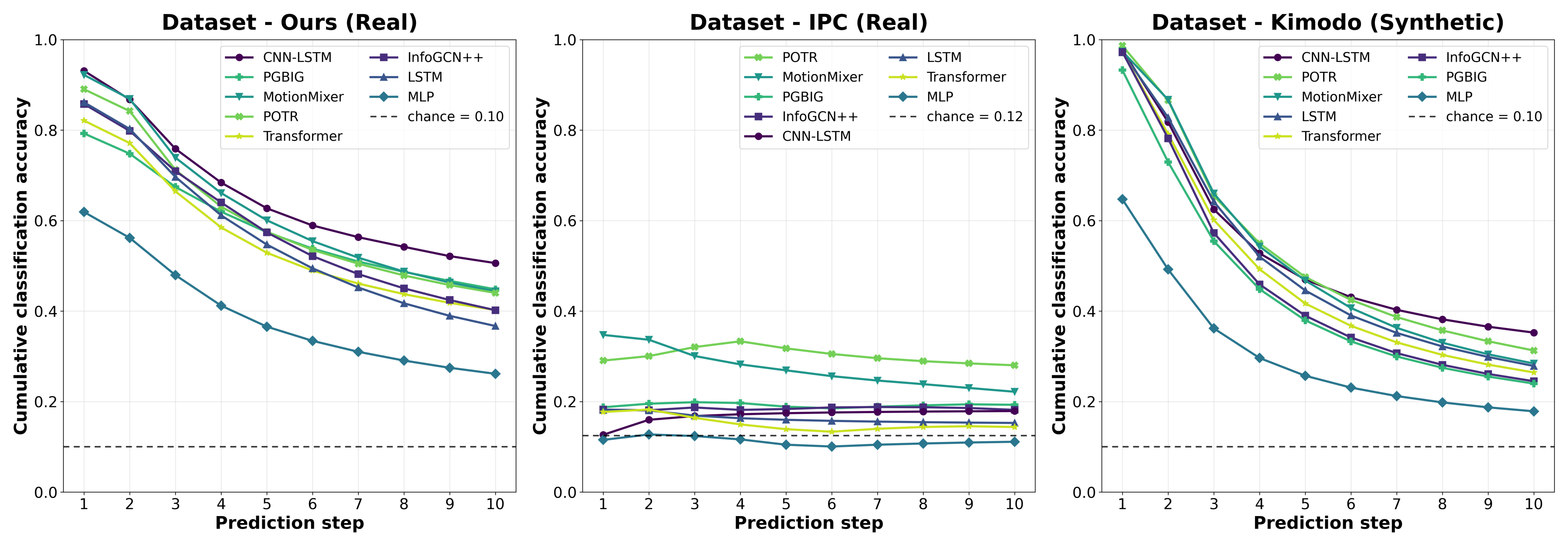}
\caption{Classification accuracy under autoregressive condition on our dataset, IPC and Kimodo (VEO3.1 and MotionLCM in the supplementary). At each step the model re-classifies a window built from its own generated motion; curves are ordered best to worst at step~10 and the dashed line is chance. Every curve starts high and declines as error accumulates; the rate of decline is what separates the models. On ours, CNN-LSTM keeps the most accuracy and the MLP the least; on Kimodo the curves start near $1.0$ and separate cleanly; on IPC they stay flat just above the $0.12$ chance line.}
\label{fig:autoreg_plot}
\end{figure}

\noindent\textbf{Autoregressive rollout stability.} Previous metrics evaluate each model on one 30-frame forecast. But an anticipatory communication system
does not run once, but rather it runs continuously, consuming a stream of motion in which
its own predictions inevitably influence the window it next reasons over. We therefore subject every trajectory model to a harder test. We feed each model
its own prediction, append it to the observation history, and re-predict, generating a self-driven rollout of 360 frames. At each of ten autoregressive
steps we re-classify the current window and record the running cumulative accuracy $\mathrm{Acc}_\ell = \frac{1}{\ell}\sum_{m\le\ell}\mathbb{1}[\hat
y^{(m)}=y]$, averaging first within each intent class and then across classes so the curve is not biased by class frequency (see Figure~\ref{fig:autoreg_plot}). The
question is no longer ``how accurate is one forecast'' but ``does the model
remain a coherent generator of in-class motion when forced to consume its own
output?''

On every dataset with recoverable signal, all curves start high and decay monotonically over the ten steps. This is the expected consequence of the autoregressive rollout (small per-step errors compound, the generated pose drifts off the data manifold, and the classifier, which now is reading an increasingly unrealistic window, loses confidence). What discriminates
the models is the rate of decay. A model that genuinely internalized the motion manifold continues to generate in-class motion and decays gracefully. However, a model that merely fit the 30-frame horizon drifts quickly toward chance. This rate is a dynamics-stability property that no single-shot metric exposes.

On our dataset the curves fan out into a clear ordering. CNN-LSTM and InfoGCN++ remain highest and decay most gracefully, holding above 0.50 and 0.40
respectively after ten self-ingesting steps, while
the MLP, having started lowest, drifts fastest and ends near 0.26. On IPC all curves sit in a narrow band just above the 0.12 chance line and stay
roughly flat across the rollout, because there is little to decay from. The models never recovered genuine recognition signal on IPC in the first place. However, looking at Kimodo, we learn that a near-perfect one-shot accuracy does not guarantee rollout stability (where one-shot recognition approaches 100\%,
self-ingesting eventually pushes most models well down toward 0.2--0.4 by the 10th step).

\noindent\textbf{The $c_{10}$ self-consistency score as an unsupervised reliability signal} The preceding experiments characterize average behavior, but neither tells a deployed system the one thing it most needs
to know (on this particular input, right now, can I trust the prediction? For an anticipatory communication
system, where acting on a misread intent is the costly failure, what is required is not higher average accuracy but a per-instance signal that flags when a prediction should be deferred. This section asks whether the model's own rollout supplies such a signal, with no access to ground truth.

\begin{figure}[htbp]
\centering
\includegraphics[width=\textwidth]{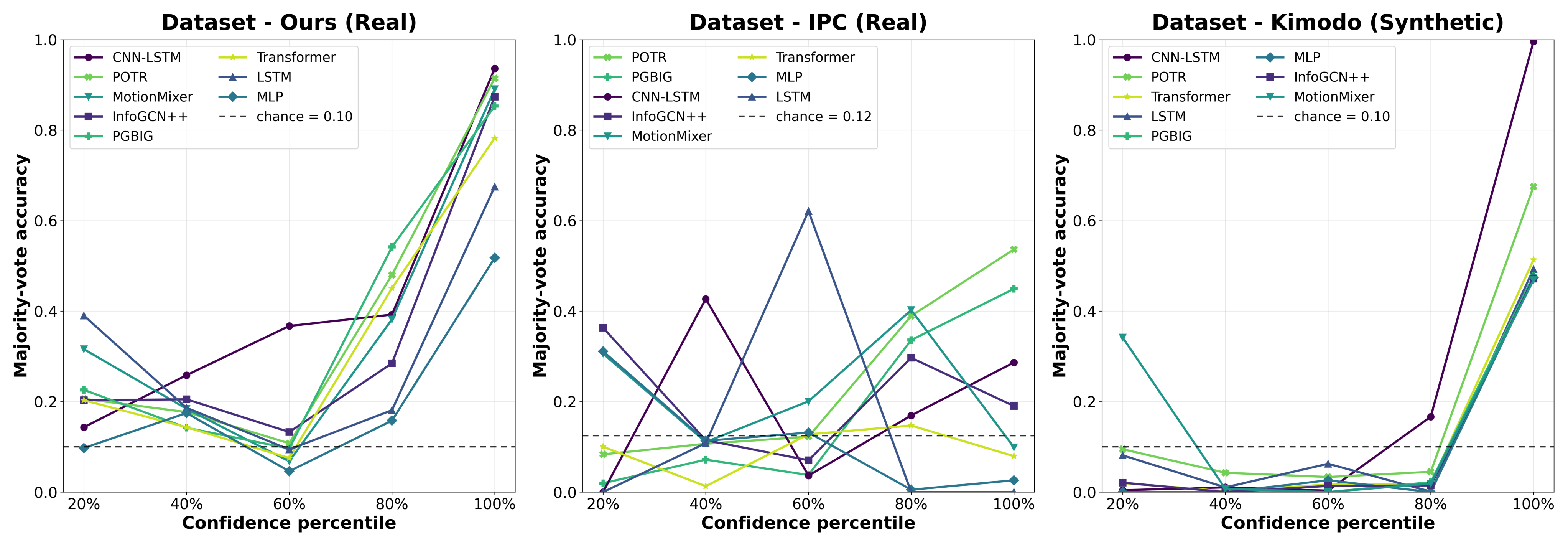}
\caption{Accuracy by $\cten$ percentile on our dataset, IPC and Kimodo (VEO3.1 and
  MotionLCM in the supplementary). Clips are sorted by $\cten$ ($100\%$ is the
  most consistent bin) and we plot majority-vote accuracy in each bin; the dashed
  line is chance. On learnable data accuracy rises with $\cten$ and the top bin
  approaches $1.0$, with Kimodo showing the sharpest rise; on IPC the curve is
  flat and irregular around chance, the case predicted by
  Equation~\eqref{eq:limit}.}
\label{fig:c10_plot}
%\vspace{-2em}
\end{figure}

For each clip we take the model's predicted class at each of the first ten
rollout steps, $\hat y^{(1)},\dots,\hat y^{(10)}$, form the majority vote $\hat
y^\star = \operatorname{mode}(\hat y^{(1)},\dots,\hat y^{(10)})$, and define the
confidence as the fraction of steps that agree with it,
\begin{equation}
c_{10} = \frac{1}{10}\sum_{m=1}^{10}\mathbb{1}\!\left[\hat y^{(m)} = \hat
y^\star\right] \in [0.1,\,1.0].
\label{eq:c10}
\end{equation}
Crucially, $c_{10}$ requires no labels (it measures only whether the model keeps voting for the same class as it consumes its own predictions) so it is available
at inference time. The hypothesis under test is that $c_{10}$ is
\emph{calibrated}, meaning that clips on which a model is internally self-consistent are
clips on which it is more likely correct. We test this by sorting each model's
test clips by $c_{10}$, splitting into five equal bins, and reporting the
majority-vote accuracy within each (see Figure ~\ref{fig:c10_plot}). A curve that lifts
toward the high-confidence bins indicates a usable reliability signal.

Across all four datasets on which recognition is learnable (\emph{Ours}, MotionLCM, VEO3.1, Kimodo), the defining feature of every curve is a pronounced
lift in the top percentile. The highest-$c_{10}$ clips reach majority-vote
accuracies of roughly 0.65--1.0, far above each model's bin-averaged accuracy and
far above chance. On \emph{Ours}, the strongest models climb to 0.85--0.93 in the
100\% bin, whilst on the clean synthetic sets the top bin reaches ${\sim}1.0$. This is
the practically important because thresholding on $c_{10}$ and acting only on
the high-consistency clips trades coverage for near-certain precision, which is
exactly the operating mode an anticipatory controller requires.

\begin{wrapfigure}{r}{0.6\textwidth}
    \centering
    %\vspace{-4.5em}
    \includegraphics[width=0.58\textwidth]{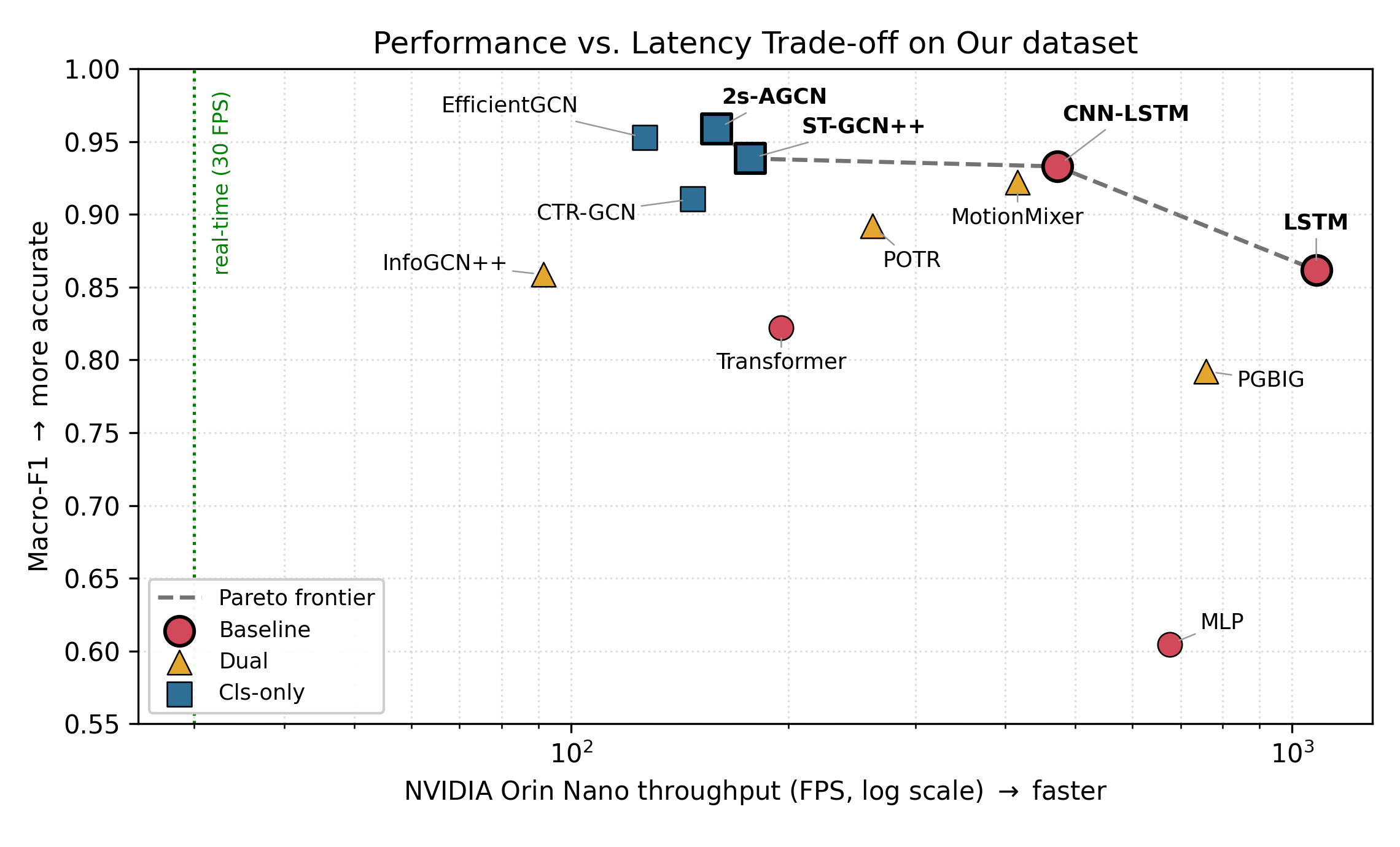}
    %\vspace{-1em}
    \caption{Accuracy--latency trade-off on the NVIDIA Orin Nano. Recognition macro-F1 versus measured throughput (FP32) for all twelve models; marker shape encodes model family, the dashed
  line is the Pareto frontier, and the green dotted line marks the 30\,FPS
  real-time threshold.}
    \label{fig:pareto_plot}
\end{wrapfigure}

We also see that calibration fails exactly where recognition fails (and this is the
experiment's most important honesty). On the IPC dataset, the curves are flat and
noisy around the 0.12 chance line, with several models wandering non-monotonically
and no reliable top-bin lift. This is precisely the failure mode a skeptical
reader should worry about, and it is the one raised against consistency-based
confidence in general. A model with no genuine recognition signal can nonetheless
be \emph{confidently self-consistent}, repeatedly voting for the same wrong class,
producing a high $c_{10}$ that is not calibrated to correctness. IPC shows what
that looks like. We report it deliberately rather than suppress it, because it
delimits the claim exactly: \textbf{$c_{10}$} is a trustworthy reliability signal
only when the underlying task is learnable from the input. When the signal is
present (the four learnable datasets), self-consistency tracks correctness at the
top; when it is absent (IPC), self-consistency decouples from correctness
entirely.

The results from Figure~\ref{fig:autoreg_plot} and Figure~\ref{fig:c10_plot} are complementary views of the same rollout. Figure~\ref{fig:autoreg_plot} asks how fast accuracy decays on average, whilst Figure~\ref{fig:c10_plot} asks, per clip, whether the model's internal agreement predicts that clip's correctness. Together they separate two
distinct questions a deployer must answer: (1) how stable is the model in general
and (2) which individual predictions can be acted upon.

\noindent\textbf{The performance--latency trade-off on edge hardware.} Previous results describe how good each model is along four axes (recognition, forecasting, rollout stability and confidence). A deployment decision requires one further axis that none of them captures: \emph{what does that quality cost at inference time on the target hardware?} An anticipatory pipeline on an NVIDIA Jetson Orin Nano has a fixed
compute budget, so the operative question is not ``which model is most
accurate'' but ``which model delivers the best accuracy attainable in real time
on this chip.'' We answer it with a Pareto analysis over all 12 models (see Figure~\ref{fig:pareto_plot}), where we measured Orin Nano throughput (frames per
second, log scale, FP32) against test-split macro-F1 on our dataset. We use macro-F1 so that
every intent counts equally regardless of clip frequency. Both axes are drawn from existing
artifacts (latency from the runtime benchmark -- Table~\ref{tab:speed_orin} and performance metrics reported in Figure~\ref{fig:class_perf}). The deployment analysis contributes three points: (1)~On the embedded platform every benchmarked model runs well above real-time, so the live trade-off for anticipatory body-pose communication is accuracy, not latency. (2)~The recognition-only frontier reduces to four non-dominated models (2s-AGCN and
ST-GCN++ for accuracy, CNN-LSTM for efficiency, LSTM for throughput), providing a
concrete, hardware-grounded recommendation. (3)~Dual-task models are dominated for recognition alone. Exactly as expected, their value lies in the
forecasting and reliability axes which the classification-only models cannot address, so the latency cost of the
joint design is justified by capability the frontier does not measure rather than
contradicted by it.

\begin{table}[t]
  \centering
  \caption{Inference performance on Jetson Orin Nano (MAXN\_SUPER power profile, PyTorch 2.5 + CUDA 12.6, FP32, batch 1).
  Latency over 50 warmup + 200 runs (mean $\pm$ std).}
  \label{tab:speed_orin}
  \setlength{\tabcolsep}{4pt}
  \renewcommand{\arraystretch}{1.1}
  \small
  \begin{tabular}{c l r r r r r}
  \toprule
   & Model & Params & FP32 & FLOPs & MACs & Latency (ms) \\
   &       & (M)    & (MB) & (M)   & (M)  & mean ($\pm$std) \\
  \midrule
  \multirow{8}{*}{\rotatebox[origin=c]{90}{\shortstack[c]{\textbf{Dual task}\\classification + generation}}}
   & CNN\_LSTM    & 0.60 &  2.30 &  55.8 &  27.9 &  2.119 ($\pm$0.054) \\
   & LSTM         & 0.36 &  1.36 &  26.4 &  13.2 &  0.926 ($\pm$0.022) \\
   & MLP          & 2.92 & 11.18 &   5.8 &   2.9 &  1.479 ($\pm$0.046) \\
   & Transformer  & 3.46 & 13.29 & 254.2 & 127.1 &  5.113 ($\pm$0.131) \\
   & MotionMixer  & 0.20 &  0.78 &  12.4 &   6.2 &  2.404 ($\pm$0.072) \\
   & POTR         & 0.39 &  1.50 &  23.0 &  11.5 &  3.824 ($\pm$0.114) \\
   & InfoGCN++    & 0.16 &  0.60 &  16.5 &   8.2 & 10.925 ($\pm$0.197) \\
   & PGBIG        & 0.72 &  2.74 &   1.4 &   0.7 &  1.317 ($\pm$0.026) \\
  \midrule
  \multirow{4}{*}{\rotatebox[origin=c]{90}{\shortstack[c]{\textbf{Single task}\\classification}}}
   & CTR\_GCN     & 0.06 &  0.25 & 114.7 &  57.4 &  6.787 ($\pm$0.157) \\
   & EfficientGCN & 0.03 &  0.16 &  30.4 &  15.2 &  7.906 ($\pm$0.190) \\
   & ST\_GCN++    & 0.11 &  0.48 & 133.1 &  66.5 &  5.646 ($\pm$0.158) \\
   & AGCN         & 0.07 &  0.29 & 122.8 &  61.4 &  6.288 ($\pm$0.156) \\
  \bottomrule
  \end{tabular}
  \vspace{-1.5em}
  \end{table}

\vspace{-1em}
\section{Closing notes}
  \label{sec:conclusions}

\iffalse
  \noindent\textbf{Conclusions.} We studied the recognition and predition of communicative intent from 2D body pose alone, a setting motivated by real-time, on-device, signal-denied human--robot
  interaction. We release a dataset with dynamic body movements, suitable for long-range detections and benchmark it against
  four further corpora chosen to span a controlled difficulty axis. Across 12 models trained under the same conditions, three findings hold. First,
  communicative intent is recoverable from body pose alone on the target distribution. Second, on the Orin
  Nano every model exceeds real-time by a wide margin, so accuracy rather than latency is the operative trade-off, and the recognition frontier reduces to four
  non-dominated models. Third, a model's autoregressive self-consistency provides a
  label-free reliability signal whose top-confidence predictions are near-certain on
  learnable data. We bound the probability that a self-consistent prediction is
  correct, show it grows with consistency, and show both analytically and
  empirically that it decouples from correctness exactly when the task is not
  learnable from the input.
\fi

\noindent\textbf{Conclusions.} We studied the recognition and prediction of communicative intent from 2D body pose alone, motivated by real-time, on-device, signal-constraint human-robot interaction. We release a dataset of large, dynamic body movements suited to long-range observation, and benchmark it against four corpora spanning a controlled difficulty axis. Across 12 models trained under the same conditions, we demonstrated that: (1) communicative intent is recoverable from body pose alone, and the difficulty axis shows the diversity that separates our real data from stereotyped synthetic motion is what makes a dataset hard to forecast. Second, on the Orin Nano every model clears real-time by a wide margin, so accuracy, not latency, is the operative constraint. Third, a model's own autoregressive self-consistency is a label-free reliability signal whose high-confidence predictions are near-certain on learnable data; we bound the probability that a self-consistent prediction is correct, show it grows with consistency, and show it decouples from correctness exactly when the task is not learnable, turning its own failure into a diagnostic for when not to trust it. Together these show body pose alone is a viable channel for communicative intent under deployment constraints, and that a system can know when to act and when to defer.

 \noindent\textbf{Reliability beyond accuracy.} To check that self-consistency is a genuine reliability signal, the supplementary material places it alongside the established toolbox for trustworthy prediction: selective-prediction risk
  (AURC)~\cite{geifman2019bias}, calibration error (ECE)~\cite{guo2017calibration}, and MC-dropout uncertainty~\cite{gal2016dropout}, among others, summarised by our $\cten$
  criterion. Across these measures the consistency-ranked predictions track the same ordering of reliability, confirming that self-consistency is
  complementary to, and not a re-description of, the usual confidence measures.

  \noindent\textbf{Toward real-world deployment.} As a first step out of the benchmark, the supplementary material reports a small real-world study run under two protocols. In the first,
  synthetic characters are composited into real scenes and driven to copy motion captured from real performers, isolating appearance and background while holding the motion fixed. In
  the second, real subjects perform the ten intents at a wide range of camera distances, stressing the full perception pipeline as keypoints degrade with scale. The trends in both
  protocols are consistent with our in-distribution conclusions, motivating a full field evaluation.

%=========================================================================
% SUPPLEMENTARY MATERIAL (from bmvc_supplementary.tex), included as appendices.
%=========================================================================
%\clearpage
\appendix

%\begin{center}
%{\Large\bf Supplementary Material}
%\end{center}
%\vspace{1em}

%-------------------------------------------------------------------------
\section{Additional details}

We present additional details about label semantics for both our dataset and IPC, that due to restricted page limit could not fit in the main submission, but are important in order to understand the reasoning behind our choice of classes/intents and also highlighting the main differences w.r.t our own selection from the IPC dataset. Also, we include in the supplementary the confidence percentile proof and additional confidence metrics and compare them against our $\cten$  proposal. We also show preliminary results on testing in a real-world scenario/environment using real drone videos. \\

\vspace{-1em}
\subsection{Intent and attribute label definitions}

For completeness we define the semantics of every class used in our experiments:
the ten communicative intents in \emph{Ours} and the eight interpersonal-circumplex
attribute codes carried by IPC~\cite{agrawal2025seamless}. These definitions are intended to make the
recognition and confusion results interpretable, and in particular to clarify
which classes are semantically adjacent.

\paragraph{The ten communicative intents (Ours)} Each clip in our dataset is labelled with the intent the actor was prompted to
convey through whole-body motion, with no reliance on facial expression or voice.
The ten intents were selected to extend the small, fixed command vocabularies of
prior drone- and robot-gesture interfaces toward a richer set of communicative
states plausibly useful in UAV/UGV interaction. Also these communication intents rather convey the natural body movement of individuals, rather than preset body gestures that one should learn or know about in order to interact with other systems.

\begin{itemize}
  \item\textbf{Affection} -- A warm, inviting disposition conveyed through soft, inward or
  embracing body movement (e.g.\ open arms drawing inward, a gentle lean toward
  the addressee). Communicates positive regard rather than a directive.
  \item\textbf{Come to me} -- A directive beckoning the addressee to approach, conveyed
  through large recruiting arm motions sweeping toward the body. Directional and
  movement-defined; its meaning is carried by the trajectory toward the self.
  \item\textbf{Dancing} -- Rhythmic, sustained whole-body motion with no directive content.
  Included as a high-energy, kinematically distinctive intent that is easy to
  separate and thus anchors the upper end of recognisability.
  \item\textbf{Disapproval} -- A negative or rejecting signal conveyed through guarded or
  warding-off motion (e.g.\ crossed arms, a halting or pushing-away gesture).
  Low-energy and posturally subtle, which makes it one of the harder intents.
  \item\textbf{Enthusiastic} -- High-energy, expansive, upward body motion signalling
  excitement or eager agreement. Semantically adjacent to \emph{happy to see you}
  and \emph{affection}, with which it is most often confused.
  \item\textbf{Go away} -- A directive dismissing or repelling the addressee, conveyed
  through shooing or pushing-outward motion away from the body. The directional
  counterpart of \emph{come to me}; the two are separable mainly by motion
  direction, which 2D projection can obscure.
  \item\textbf{Happy to see you} -- A greeting conveying positive recognition (e.g.\ waving,
  raised welcoming arms). Low-to-moderate energy and expression-driven, hence
  adjacent to \emph{enthusiastic} and \emph{affection}.
  \item\textbf{Laughing} -- Amusement conveyed through characteristic torso and shoulder
  motion. Kinematically distinctive and among the most reliably recognised intents.
  \item\textbf{Surprise} - A sudden, short-onset reaction conveyed through a rapid startle
  or recoil of the upper body. Defined by its temporal sharpness, which is why a
  time-flattening model recovers it poorly.
  \item\textbf{We have a deal} - An agreement gesture (e.g.\ a handshake-like
  or sealing motion). Its body-only realization is comparatively subtle, which is
  reflected in its lower recall across models.
\end{itemize}

Several intents are deliberately close in affective content: \emph{affection},
\emph{enthusiastic}, \emph{happy to see you} form a warm-positive cluster, and
\emph{come to me}/\emph{go away} form a directional pair distinguished mainly by
motion sense. The recurring cross-model confusions among exactly these classes
(see the Experiments section in the main submission) therefore reflect genuine semantic and projective
ambiguity rather than a defect of any single architecture.

\paragraph{The eight IPC attribute codes}
The IPC labels originate from Meta's Seamless Interaction dataset~\cite{agrawal2025seamless}, whose annotations are grounded in the \emph{interpersonal
circumplex}, a standard model in personality and social psychology
\cite{wiggins1979psychology,locke2011circumplex}. The circumplex organises interpersonal stance along
two orthogonal axes: \textbf{agency} (assertiveness / dominance) and
\textbf{communion} (warmth / affiliation). Each code has the form
$\texttt{A}\langle a\rangle\texttt{C}\langle c\rangle$, where the letter
$\langle a\rangle$ grades the \emph{assertiveness} level and the letter
$\langle c\rangle$ grades the \emph{warmth} level, each drawn from
$\{\texttt{M},\texttt{N},\texttt{P}\}$. All eight codes lie on the assertive side
of the agency axis and differ in the degree of assertiveness and in warmth.
We use the mapping from the originating annotation scheme:

\begin{description}
  \item[\texttt{AMCM}] - Moderately assertive, balanced warmth.
  \item[\texttt{AMCN}] Moderately assertive, neutral warmth.
  \item[\texttt{AMCP}] Moderately assertive, warm.
  \item[\texttt{ANCM}] Assertive, cold.
  \item[\texttt{ANCP}] Assertive, warm (dominant--friendly).
  \item[\texttt{APCM}] Highly assertive, moderately cold.
  \item[\texttt{APCN}] Highly assertive, neutral warmth.
  \item[\texttt{APCP}] Highly assertive, warm.
\end{description}

\noindent Read together, the assertiveness position $\langle a\rangle$ increases in
the order $\texttt{M}\!\rightarrow\!\texttt{N}\!\rightarrow\!\texttt{P}$
(moderately assertive, assertive, highly assertive), while the warmth position
$\langle c\rangle$ ranges from cold/moderate through neutral to warm. The eight
codes thus tile a graded region of the circumplex rather than naming eight
distinct gestures.

Two properties of this label space explain its behaviour as a negative control in
our experiments. First, the codes are \emph{attributes of interpersonal stance}
rather than discrete communicative gestures, so they need not correspond to any
distinctive whole-body motion. Second, the scheme is \emph{graded along two
continuous axes}: neighbouring codes differ only by one level on one axis (e.g.\
\texttt{AMCM}/\texttt{AMCN}/\texttt{AMCP} share an assertiveness level and step
through warmth; \texttt{AMCx}/\texttt{ANCx}/\texttt{APCx} share a warmth level and
step through assertiveness), so even a perfect annotator would expect substantial
overlap between adjacent codes. Together these explain why IPC stance is largely
unrecoverable from gross 2D body motion alone: the
signal is encoded substantially in the face and in fine interactive behaviour, and
the label geometry is graded rather than categorical.

\subsection{Proposition proof}
 We present the proof for proposition 1.
Let $C$ denote the correct class and $W$ the distractor (wrong) class. We assume without loss of generality that there is a single main distractor class and that the probability of generating samples from any other incorrect class is negligible. This reduces the analysis to a binary classification setting; the same reasoning extends naturally to the multi-class case.

We wish to estimate the probability of predicting the correct class on the original input, given $n$ successive predictions consistent with the first one. By Bayes' rule:

\begin{equation}
    P(C \mid n) = \frac{P(n \mid C)\,P(C)}{P(n \mid C)\,P(C) + P(n \mid W)\,P(W)},
\end{equation}

where $n$ denotes the consistency count. Using the notation $f_r, f_g, P_r, P_g$ introduced earlier:

\begin{itemize}
    \item $P(n \mid C)\,P(C)$: the probability of obtaining $n$ correct recognitions and $n-1$ correct self-generations.
    \item $P(n \mid W)\,P(W)$: the probability of obtaining $n$ wrong recognitions with $n-1$ in-class generations, plus the probability of a wrong recognition followed by a wrong distractor generation followed by $n-1$ consistent recognitions and $n-2$ correct self-generations with respect to the distractor.
\end{itemize}

Substituting $P(C) = f_r \cdot P(W)$ and $P_g(C) = f_g \cdot P_g(W)$, with $f_r, f_g > 1$, we obtain:

\begin{equation}
    P(C \mid n)
    = \frac{f_r^n f_g^{n-1}}{f_r^n f_g^{n-1} + f_g^{n-1} + f_r^{n-1} f_g^{n-2}}
    = \frac{1}{1 + \dfrac{1}{f_r^n} + \dfrac{1}{f_r f_g}}.
    \label{eq:pcorrect}
\end{equation}

Since $f_r > 1$, the term $1/f_r^n$ decreases monotonically with $n$, so $P(C \mid n)$ is strictly increasing in $n$. As $n \to \infty$, it converges to the upper bound:

\begin{equation}
    \lim_{n \to \infty} P(C \mid n) = \frac{1}{1 + \dfrac{1}{f_r f_g}}.
\end{equation}

When the distractor is easily confused with the correct class ($f_r, f_g \approx 1$), this upper bound approaches $1/2$, reflecting near-chance accuracy regardless of consistency. Conversely, when $f_r$ and $f_g$ are large --- meaning the correct class is clearly distinguishable --- the bound approaches $1$. In all cases, higher consistency $n$ is strictly correlated with higher classification accuracy.

\subsection{Qualitative results}

\begin{figure}[htbp]
\centering
\includegraphics[width=\textwidth]{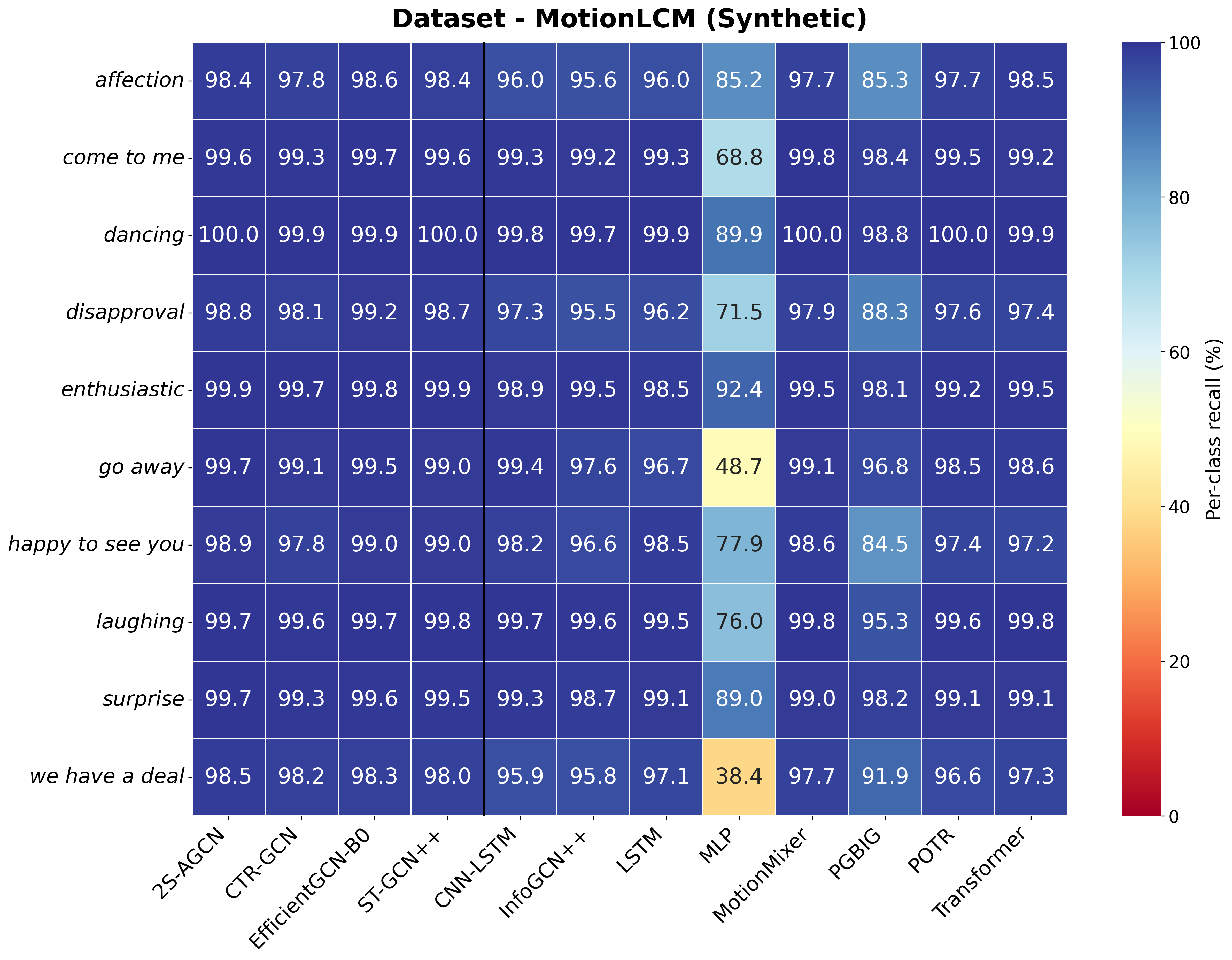}
\caption{Per-class recall (\%) on the synthetic MotionLCM dataset. Results on the other datasets are presented in the main submission. Rows are intents. The methods run left to right as the four classification-only graph recognizers (2s-AGCN, CTR-GCN, EfficientGCN-B0, ST-GCN++) followed by the eight dual-task models. Blue is high recall, red is low.}
\label{fig:motionlcm_class_perf}
\end{figure}

MotionLCM is the lowest-variance generator in our suite, and its panel (see Figure~\ref{fig:motionlcm_class_perf}) is
correspondingly the most saturated: every sequence-aware model attains 95--100\%
recall on essentially every class. The only structured departure is the MLP
column, which degrades on the temporally defined intents (\emph{we have a deal}
38.4\%, \emph{go away} 48.7\%, \emph{happy to see you} 77.9\%) while remaining
strong on the more distinct ones. We stress that this panel is a
\emph{control}, not a performance result: high intra-distribution recall on a
low-variance synthetic distribution is expected almost by construction, and we
use it only to certify that the recognition pipeline saturates when class signal
is unambiguous---thereby attributing the failures seen on \emph{Ours},
\emph{VEO3.1} and \emph{IPC} to data and class properties rather than to
implementation. Read alongside Kimodo (balanced classes, also near-ceiling) and
VEO3.1 (high-variance, strongly discriminative), MotionLCM completes a synthetic
difficulty axis that brackets the real datasets from below.

\begin{figure}[htbp]
\centering
\includegraphics[width=\textwidth]{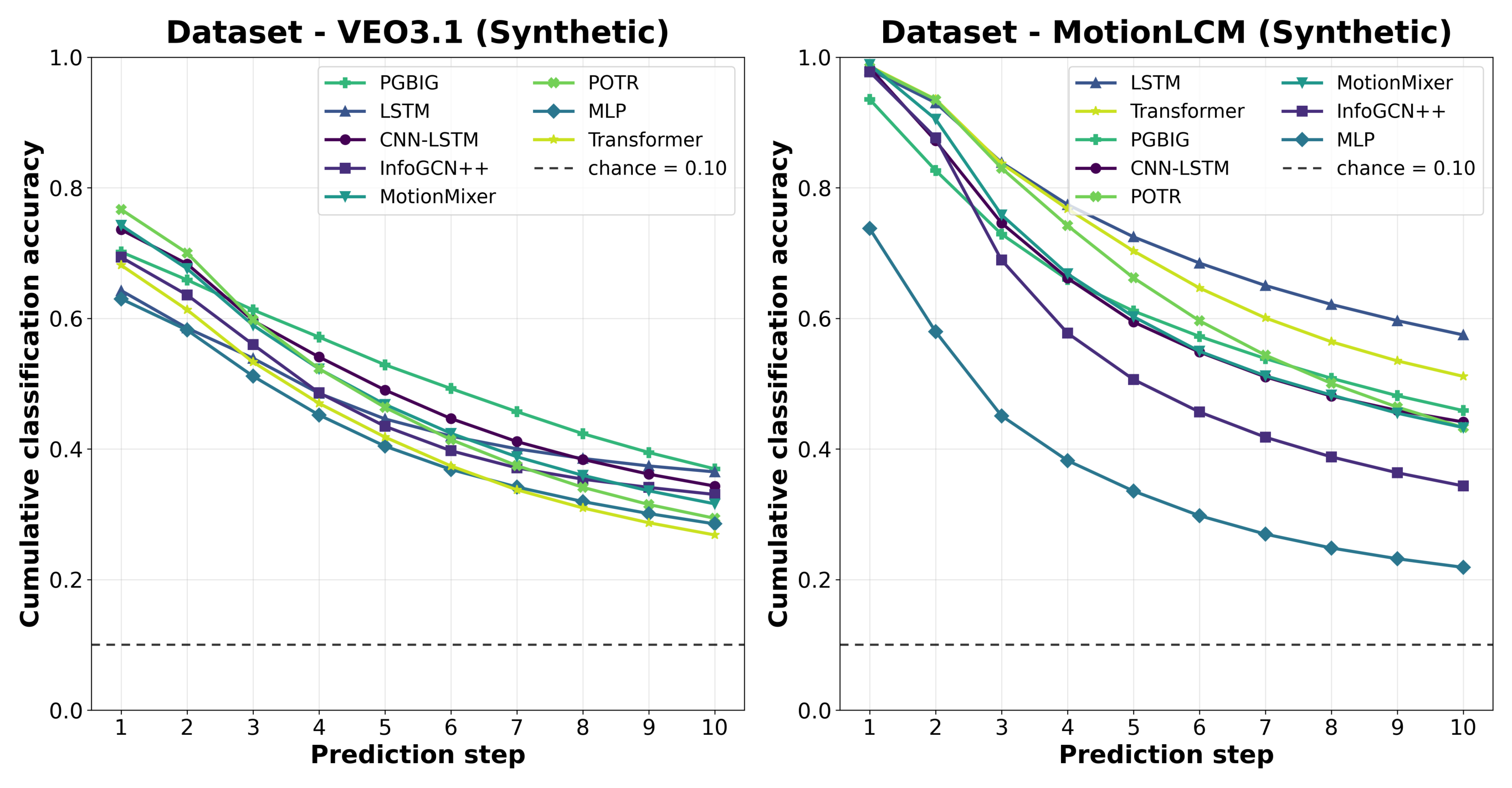}
\caption{Classification accuracy under autoregressive condition on the synthetic VEO3.1 and MotionLCM datasets. At each step the model re-classifies a window built from its own generated motion; curves are ordered best to worst at step~10 and the dashed line is chance. Every curve starts high and declines as error accumulates; the rate of decline is what separates the models.}
\label{fig:autoreg_supp}
\end{figure}

As the most difficult synthetic distribution, VEO3.1 (see Figure~\ref{fig:autoreg_supp}) produces lower starting accuracies
(${\approx}0.63$--$0.77$, consistent with its harder one-shot recognition (results presented in main submission) that decay into the 0.27--0.37 band by step ten. The relative ordering of models is consistent with the other panels (PGBIG and the
stronger sequence models retain the most, the MLP and Transformer the
least) reinforcing that the stability ranking is a model property that transfers
across generation sources of very different appearance and noise characteristics. For MotionLCM, The LSTM (top curve, ${\sim}0.57$ at step ten) and
Transformer retain substantially more accuracy under self-ingesting than InfoGCN++
(${\sim}0.34$) and the MLP (${\sim}0.22$), despite InfoGCN++ holding a single-shot
forecasting-error advantage. We read this as the
clearest available demonstration that position-error minimization and rollout
stability are distinct objectives, and that the former can be achieved by
over-smoothing at the cost of the latter.

\begin{figure}[htbp]
\centering
\includegraphics[width=\textwidth]{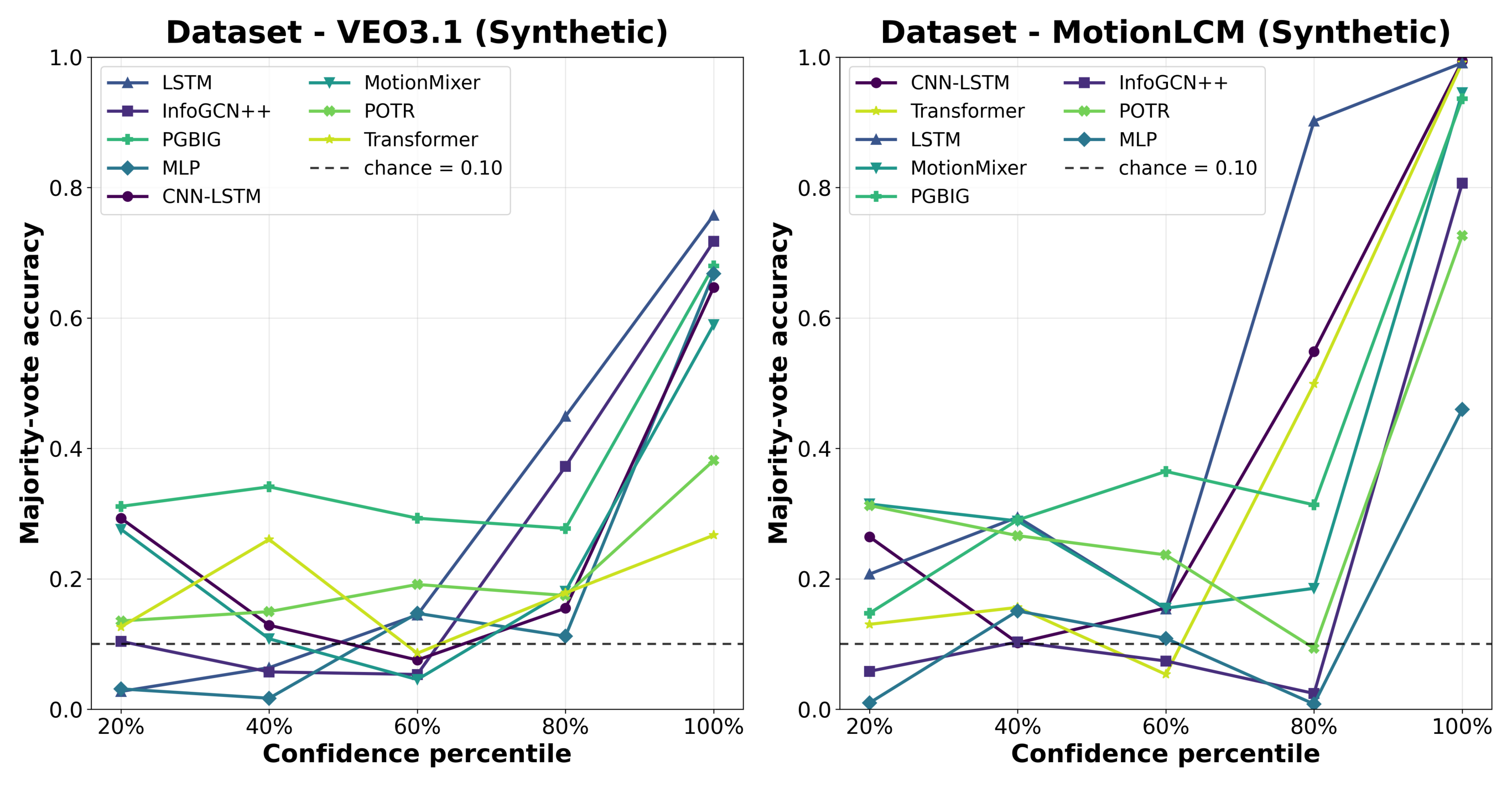}
\caption{Accuracy by $\cten$ percentile on the VEO3.1 and
  MotionLCM datasets. \label{fig:c10_supp}}
\end{figure}

Figure~\ref{fig:c10_supp} shows that on VEO3.1 the top-percentile accuracies reach ${\sim}0.6$--$0.75$ for the stronger
sequence models (LSTM, InfoGCN++, MotionMixer, CNN-LSTM) over flatter low/middle
bins, while POTR and the Transformer remain comparatively flat (consistent with
their weaker recognition on this distribution presented in the main submission). On
MotionLCM the top-bin lift is the most dramatic of any dataset, with several
models leaping from near-chance middle bins to ${\sim}0.9$--$1.0$, while a subset
lift less, mirroring their reduced rollout retention.

\subsection{Synthetic data generation prompts}
We present additional details regarding the prompts and settings used fo the synthetic data generation.

\noindent\textbf{Kimodo} We prompt it with ``\textit{a person showing X body language}'' for each of the 10 emotions and sample $120$ motions per class with random seed $0$ to
build the training set ($N{=}1{,}200$), then sample a further
$30$~motions per class with seed $42$ to build the test set
($N{=}300$).

\noindent\textbf{MotionLCM}  We use this model to generate 10M frames, split into the same 10 emotions, prompted as \textit{a person <intent>}. The \textit{body language} resulted in similar outputs, so we kept the original prompt.

\noindent\textbf{VEO3.1} The same set of classes apply, the prompt being \textit{A person showing "<intent>" body language gesture, full body shot.} The prompt was sligthy altered from motionLCM due to empirical observations. Several classes had prompt adherence issues, i.e., "we have a deal" generally showed two persons shaking hands, but most results were similar in appearance with our dataset.

\section{Additional consistency metrics}

We compare our $\cten$ consistency metric against other label and label-free robustness methods available in the literature and show the results in Table ~\ref{tab:consistency-top4}.

\begin{figure*}[t]
    \centering
    \subfigure[Ours]{\includegraphics[width=0.32\linewidth]{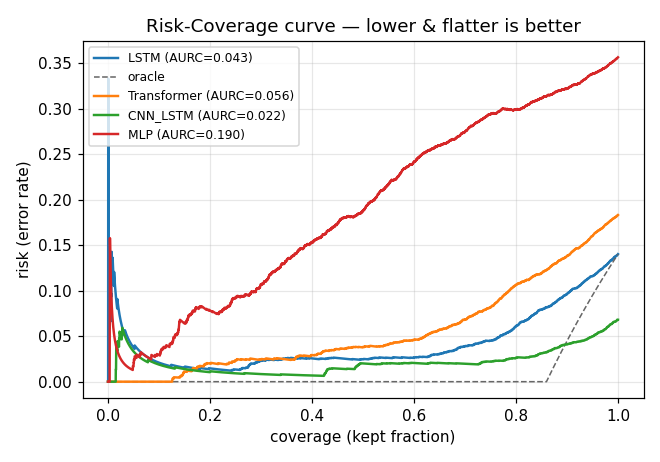}\label{fig:rc-ours}}\hfill
    \subfigure[VEO~3.1]{\includegraphics[width=0.32\linewidth]{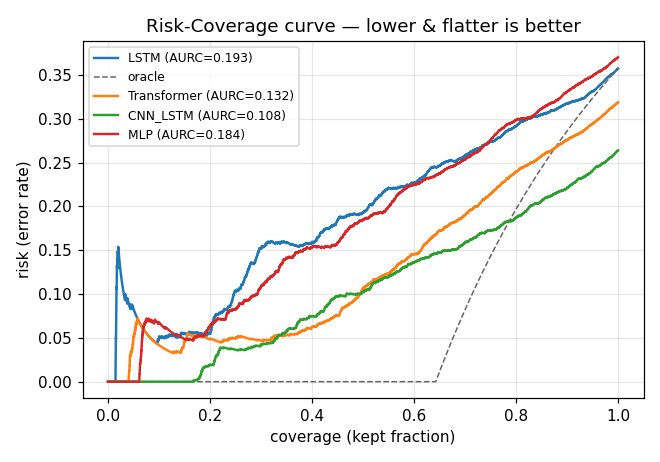}\label{fig:rc-veo3}}\hfill
    \subfigure[MotionLCM]{\includegraphics[width=0.32\linewidth]{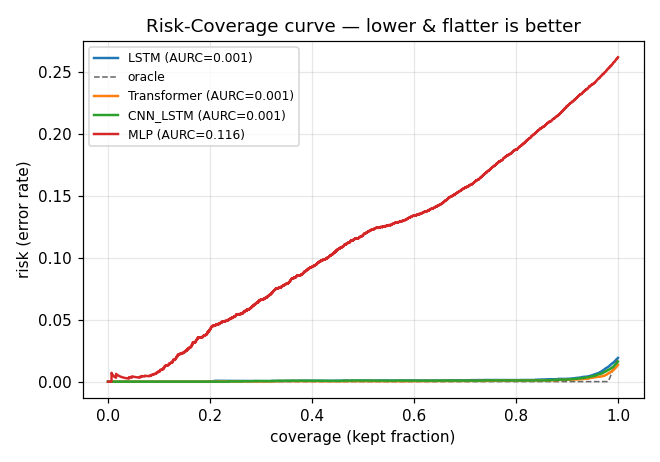}\label{fig:rc-motionlcm}}
    \vspace{5mm}
    \caption{\textbf{Risk-coverage curves} for the trained
    architectures on each test set. Coverage on the x-axis is the
    fraction of test samples retained after thresholding by softmax
    margin; risk on the y-axis is the classification error on those
    retained samples. Lower-and-left is better. AURC corresponds to the
    area under each curve and matches the AURC column of
    Tab.~\ref{tab:consistency-top4}.}
    \label{fig:risk-coverage}
  \end{figure*}

\paragraph{Discussion: predictive performance vs.\ data realism.}
The three test sets in Tab.~\ref{tab:consistency-top4} and
Fig.~\ref{fig:risk-coverage} cover the full spectrum from
fully-synthetic, low-noise skeletons (MotionLCM, diffusion-
generated trajectories) to recorded human video (Ours, $1{,}195$
real clips processed by YOLO11-pose) and a middle-ground
photorealistic-but-synthetic regime (Veo~3.1, AI-generated video
clips). Three robust patterns emerge across the four architectures we
evaluated (LSTM, Transformer, CNN\_LSTM and MLP):

\begin{itemize}
\setlength{\itemsep}{2pt}
\item \textbf{MotionLCM is not a proper benchmark.} All three
recurrent / convolutional architectures reach $\approx\!98$\,\%
accuracy with near-zero AURC ($\le\!0.001$), ECE ($\le\!0.004$) and
Brier ($\le\!0.03$). The MLP baseline trails badly ($73.8$\,\%
accuracy, AURC\,$0.116$), which on this clean distribution is more
a comment on model capacity than on data difficulty.
Selective-classification has effectively nothing to do
(Fig.~\ref{fig:rc-motionlcm}); the dataset is too easy to differentiate
the three top architectures.

\item \textbf{Ours stresses model design.} On the real-video set,
CNN\_LSTM is the strongest classifier ($93.2$\,\% accuracy,
AURC\,$0.022$, mean-margin $0.97$) and the Transformer is the worst of
the three deep architectures ($81.7$\,\%, AURC\,$0.056$). Despite the
accuracy gap, all three remain almost equally robust to perturbations
(Jitter\,$\approx\!0.97$--$0.98$, Mirror\,$\approx\!0.76$--$0.84$),
suggesting that the failures are in label
boundaries rather than in temporal stability. The risk-coverage
curves in Fig.~\ref{fig:rc-ours} make the architecture ranking
visually obvious -- CNN\_LSTM dominates the low-coverage end.

\item \textbf{VEO~3.1 is the hardest dataset overall.} Accuracies
collapse to $63$--$74$\,\% across all four models, AURC rises by
$\approx\!5\times$ compared to Ours, and calibration degrades (ECE
$0.09$--$0.15$, Brier $0.40$--$0.51$). CNN\_LSTM still leads but with a
much smaller margin, and the gap between models shrinks --- evidence
that the synthetic-but-realistic regime introduces failure modes that
all four classifiers struggle with equally. The risk-coverage curves
in Fig.~\ref{fig:rc-veo3} stay flatter and higher than on Ours,
indicating that softmax margin is a poor selector under this
distribution shift.
\end{itemize}

Two consequences for downstream use: (1) reporting model performance on
MotionLCM \emph{alone} dramatically overstates real-world reliability
($\Delta$\,accuracy\,$\approx\!15$\,pp vs.\ Ours, $\approx\!30$\,pp
vs.\ Veo~3.1), so MotionLCM set should be reserved for
sanity-checking and large-scale pre-training, not headline numbers;
and (2) the $\cten$ sliding-window self-consistency score is the only
metric that stays informative across all three datasets
($0.60$--$0.82$ range), supporting its use as a deployment-time
no-label proxy for trustworthiness.

\begin{table*}[t]
\centering
\caption{Per-cell consistency metrics on three datasets, top four
trained architectures only. Accuracy, MAE and the AURC/E-AURC
selective-classification metrics are computed against the same
$10$-class emotion taxonomy; ECE and Brier measure calibration;
Margin reports the mean and 10th-percentile softmax margin; Jitter,
Mirror and Drop measure label invariance under three perturbations
(temporal jitter, horizontal mirror, MC-dropout). MC entropy / MI
summarize MC-dropout disagreement.}
    \vspace{3mm}

\setlength{\tabcolsep}{3pt}
\renewcommand{\arraystretch}{1.08}
\resizebox{\linewidth}{!}{%
\begin{tabular}{ll rrrr rr c rrr rr r}
\toprule
\textbf{Dataset} & \textbf{Model}
 & \textbf{Acc $\uparrow$} & \textbf{MAE $\downarrow$}
 & \textbf{AURC $\downarrow$} & \textbf{E-AURC $\downarrow$}
 & \textbf{ECE $\downarrow$} & \textbf{Brier $\downarrow$}
 & \textbf{Margin (mean\,/\,p10)}
 & \textbf{Jitter $\uparrow$} & \textbf{Mirror $\uparrow$} & \textbf{Drop $\uparrow$}
 & \textbf{MC ent.\ $\downarrow$} & \textbf{MC MI $\downarrow$}
 & \textbf{C10 $\uparrow$} \\
\midrule

\multirow{4}{*}{Ours}
 & LSTM        & 0.860 & 0.063 & 0.043 & 0.033 & 0.093 & 0.233 & 0.916\,/\,0.713 & 0.976 & 0.801 & 0.810 & 0.149 & 0.000 & 0.721 \\
 & Transformer & 0.817 & 0.064 & 0.056 & 0.038 & 0.083 & 0.284 & 0.827\,/\,0.389 & 0.980 & 0.761 & 0.807 & 0.286 & 0.000 & 0.762 \\
 & CNN\_LSTM   & 0.932 & 0.065 & 0.022 & 0.019 & 0.053 & 0.124 & 0.967\,/\,0.972 & 0.971 & 0.839 & 0.473 & 0.057 & 0.000 & 0.775 \\
 & MLP         & 0.643 & 0.072 & 0.190 & 0.117 & 0.054 & 0.505 & 0.398\,/\,0.069 & 0.979 & 0.647 & 0.705 & 1.121 & 0.000 & 0.771 \\
\midrule

\multirow{4}{*}{Veo~3.1}
 & LSTM        & 0.643 & 0.073 & 0.193 & 0.120 & 0.092 & 0.506 & 0.574\,/\,0.116 & 0.975 & 0.690 & 0.778 & 0.743 & 0.000 & 0.823 \\
 & Transformer & 0.682 & 0.067 & 0.132 & 0.075 & 0.150 & 0.457 & 0.718\,/\,0.163 & 0.978 & 0.695 & 0.749 & 0.482 & 0.000 & 0.696 \\
 & CNN\_LSTM   & 0.736 & 0.075 & 0.108 & 0.069 & 0.110 & 0.402 & 0.744\,/\,0.191 & 0.976 & 0.692 & 0.629 & 0.445 & 0.000 & 0.757 \\
 & MLP         & 0.630 & 0.078 & 0.184 & 0.105 & 0.056 & 0.501 & 0.395\,/\,0.054 & 0.976 & 0.653 & 0.620 & 1.230 & 0.000 & 0.673 \\
\midrule

\multirow{4}{*}{MotionLCM}
 & LSTM        & 0.981 & 0.035 & 0.001 & 0.001 & 0.003 & 0.030 & 0.965\,/\,0.950 & 0.768 & 0.979 & 0.627 & 0.070 & 0.000 & 0.788 \\
 & Transformer & 0.986 & 0.032 & 0.001 & 0.001 & 0.003 & 0.021 & 0.977\,/\,0.982 & 0.774 & 0.987 & 0.608 & 0.038 & 0.000 & 0.779 \\
 & CNN\_LSTM   & 0.984 & 0.036 & 0.001 & 0.001 & 0.004 & 0.026 & 0.977\,/\,0.983 & 0.746 & 0.981 & 0.466 & 0.048 & 0.000 & 0.615 \\
 & MLP         & 0.738 & 0.042 & 0.116 & 0.078 & 0.165 & 0.421 & 0.364\,/\,0.060 & 0.964 & 0.906 & 0.443 & 1.175 & 0.000 & 0.602 \\

\bottomrule
\end{tabular}}

\label{tab:consistency-top4}
\end{table*}

\section{Real-world testing}
To test our model in real-world scenarios/applications, we have explored its capabilities in visual data captured from drones. We present two scenarios: (1) human placement from real motion transfer and a photorealistic 3D avatar and (2) real footage placed in a real UAV video and real UAV footage alone (participant captured from different heights).

\subsection{Real world placement}

\begin{figure*}[t]
  \centering
  \includegraphics[width=\linewidth]{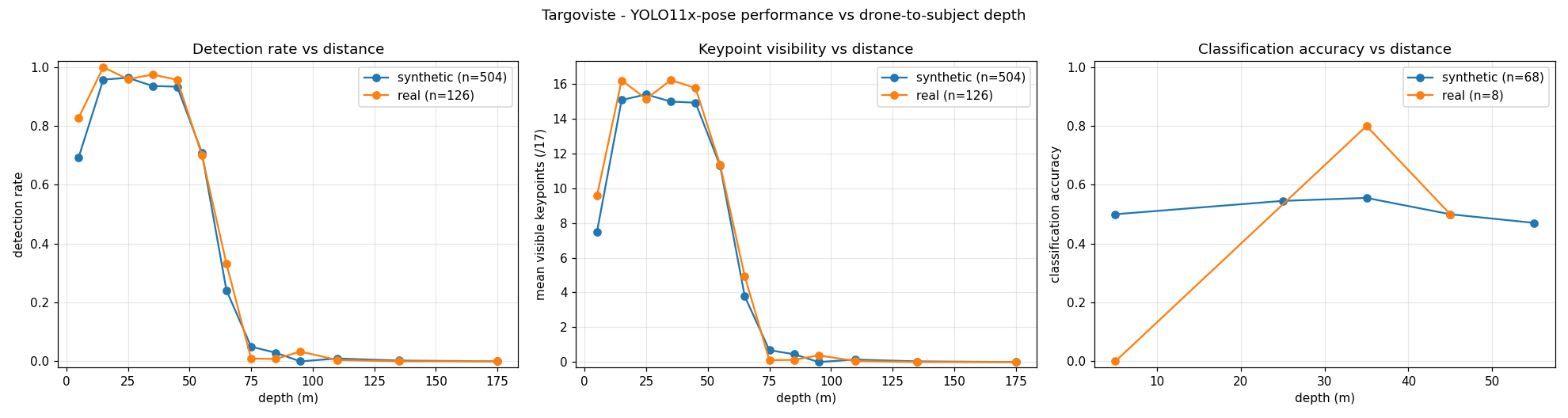}
  \caption{\textbf{Detection, keypoint visibility and classification performance vs.\
  depth on the \textsc{Park} real footage placement study.} Blue: photorealistic avatars with human-powered motions, orange: real participants blended in the scene. \emph{Left:}
  detection rate (frames with an IoU\,$\ge\,0.3$ match to the placement
  bounding box, normalized by the number of frames in which the
  placement is visible). \emph{Center:} mean number of COCO keypoints
  with detector confidence above $0.5$. Each marker aggregates all
  placements whose foot-point depth falls in the corresponding $10$\,m
  bin (last three bins widened to $20$, $30$, $50$\,m to absorb the
  long, sparsely populated tail). The two series, synthetic placements
  ($n\!=\!496$) and real placements ($n\!=\!124$), track each other
  closely. \emph{Right:} Classification performance using the transformer-based model. }
  \label{fig:park-yolo-perf}
\end{figure*}

\paragraph{Detection vs. depth on the \textsc{Park} placement
study.}
Due to the lack of necessary participants and need for a robust dataset, we transfer the body pose from actors to 3D avatars and place them in real scenes. We capture the motion with SMPL-X, render a 3D avatar, do a semantic segmentation for pavement or flat areas (low vegetation, roofs) using SAM3~\cite{carion2025sam3} and
we measure how aerial pose estimation degrades with the subject's
distance to the drone on a 252\,s 4K (\,$3840\!\times\!2160$\,) clip
captured over a park. The clip is split into
$63$~non-overlapping $120$-frame sequences; each sequence contains
$10$~placed characters ($8$ synthetic SMPL-X sprites + $2$ real
emotion-clip cutouts) tracked through the sequence with
CoTracker~\cite{karaev2024cotracker} and composited onto pavement
regions identified by SAM3~\cite{carion2025sam3}. Per-character
ground-truth depth is recovered from a MoGe-2~\cite{wang2025moge2}
monocular depth map sampled at the foot-point. We run YOLO11x-pose
\cite{jocher2024yolo11} in a $640\!\times\!640$ tiled sweep ($128$~px
overlap) over each frame, match each detection to the placement bounding
box at IoU\,$\ge\,0.3$, and record the per-character (i)~detection rate
across the frames where the placement is visible, and
(ii)~mean number of COCO keypoints whose model confidence exceeds
$0.5$. Figure~\ref{fig:park-yolo-perf} reports both metrics in
$10$\,m depth bins out to $100$\,m and wider bins thereafter.

\begin{figure*}[t]
    \centering
    \includegraphics[width=\linewidth]{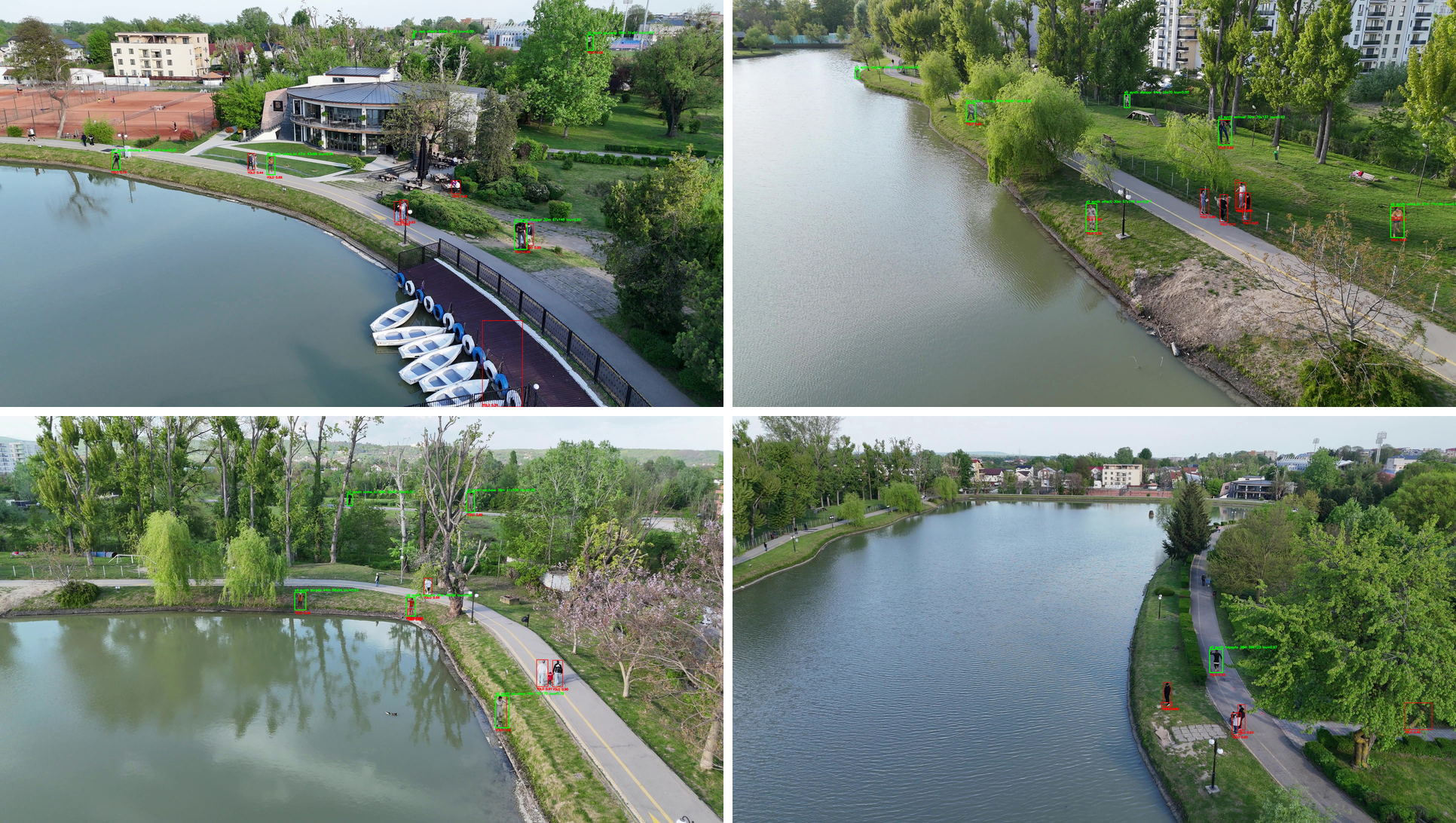}
    \vspace{1mm}
    \caption{\textbf{Samples with placement and detections from the \textsc{Park} dataset} Ground-truth (green) versus YOLO (red) detections on four aerial sequences from the \textit{Park} scene, with synthetic placements (i.e., motion capture, SMPL-X parameters, 3D avatar, depth estimation, scale-aware blending). Despite long range and small subjects, YOLO recovers the salient
  pedestrians and objects, with similar performance for both synthetic (human motion source + 3D avatar) and real (human pasted in scene). }
    \label{fig:gt_vs_yolo_park}
  \end{figure*}

\paragraph{Discussion.}
Both detection rate and keypoint visibility are essentially saturated
($\det\!\approx\!0.95$, $\sim\!15$/$17$ visible keypoints) over the
$10$--$50$\,m range, then collapse over a narrow transition around
$50$--$80$\,m, and are effectively zero beyond $100$\,m. The
$0$--$10$\,m bin shows a modest drop (especially for synthetic
placements) because subjects fill the frame and a fraction of their
keypoints fall outside the tile grid; this is a tiling artefact rather
than a model limitation. The near-identical synthetic and real curves
confirm that the renderer's photometric integration is consistent
enough that YOLO does not preferentially favour one source over the
other---synthetic placements degrade with depth in the same way the
real-clip cutouts do, validating the composite as a useful
controllable-distance benchmark for downstream classification.

\subsection{Real footage}

\paragraph{Detection vs.\ depth on the \textsc{Resort} real-person
study.}
The \textsc{Resort} set complements the \textsc{Park} placement
benchmark with raw drone footage of nine real human subjects, one per
body-language class, each captured at 4K
(\,$3840\!\times\!2160$\,)~/~$30$\,fps for $30$\,s ($900$ frames).
For every clip we track the central-most YOLO11x-pose
\cite{jocher2024yolo11} detection from frame~$0$ across the whole clip
with a scale-tolerant greedy tracker (nearest-centroid assignment with
a per-frame budget of $3\!\times\!$ the previous bounding-box
diagonal, floored at $60$\,px), so the same identity is followed as
the subject walks toward or away from the drone and the bounding box
shrinks. Each tracked trajectory is split into seven non-overlapping
$120$-frame sequences (matching the granularity of the
\textsc{Park} study) and we report per-sequence (i)~detection
rate, defined as the fraction of frames in which the central-person
track is alive, and (ii)~mean count of COCO keypoints with detector
confidence above $0.5$ in those tracked frames. Per-frame foot-point
depth is sampled from the MoGe-2~\cite{wang2025moge2} depth map.

% TODO figure hidden
\begin{figure}[t]
  \centering
  \includegraphics[width=\linewidth]{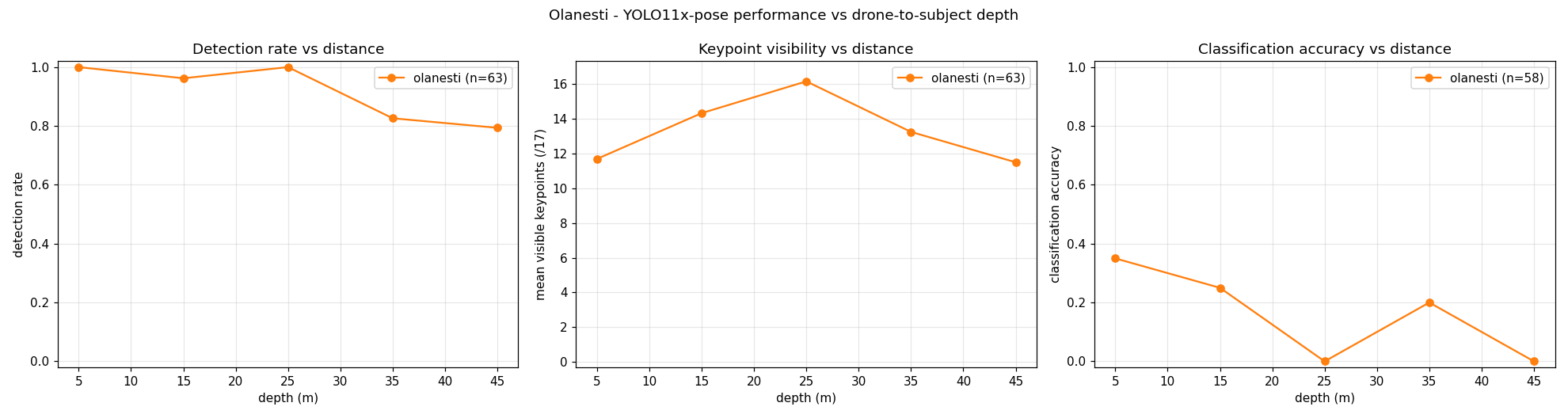}
    \vspace{1mm}
  \caption{\textbf{Detection, keypoint visibility and classification performance vs.\
  depth on the \textsc{Resort} real-person study} (one real subject
  per clip, $n\!=\!63$ sequences across nine clips). \emph{Left:}
  fraction of frames in which the central-person tracker held an
  IoU-validated YOLO detection. \emph{Center:} mean number of COCO
  keypoints with detector confidence above $0.5$. Only the
  ``real'' series is shown -- there are no synthetic placements in
  this study. \emph{Right:} Classification performance, transformer model - even though most keypoints are visible, the detection does not yield a satisfactory classification.}
  \label{fig:resort-yolo-perf}
\end{figure}

\paragraph{Discussion.}
Because every clip contains a single, deliberately framed actor, the
tracker holds the central identity in $\sim\!95$\% of frames across all
nine clips and detection rate stays above $0.85$ throughout the
populated depth range. Keypoint visibility peaks (${\sim}16$\,/\,17)
in the $20$--$30$\,m band, dips below $12$ inside the closest
($<\!10$\,m) bin -- there, the subject can fill more than a third of
the $2160$\,px image height, so a fraction of the body simply falls
outside the tile grid -- and degrades smoothly past $40$\,m as the
silhouette shrinks. The narrow operating range
($5$--$65$\,m, no clip ever further than $66$\,m) constrains how
informative this benchmark is at long distances; for the long-range
($>\!100$\,m) regime see the matching \textsc{Park} placement
results in Fig.~\ref{fig:park-yolo-perf}, where the same
detector falls off a cliff between $50$\,m and $80$\,m.

\begin{figure}[t]
    \centering
    \includegraphics[width=\linewidth]{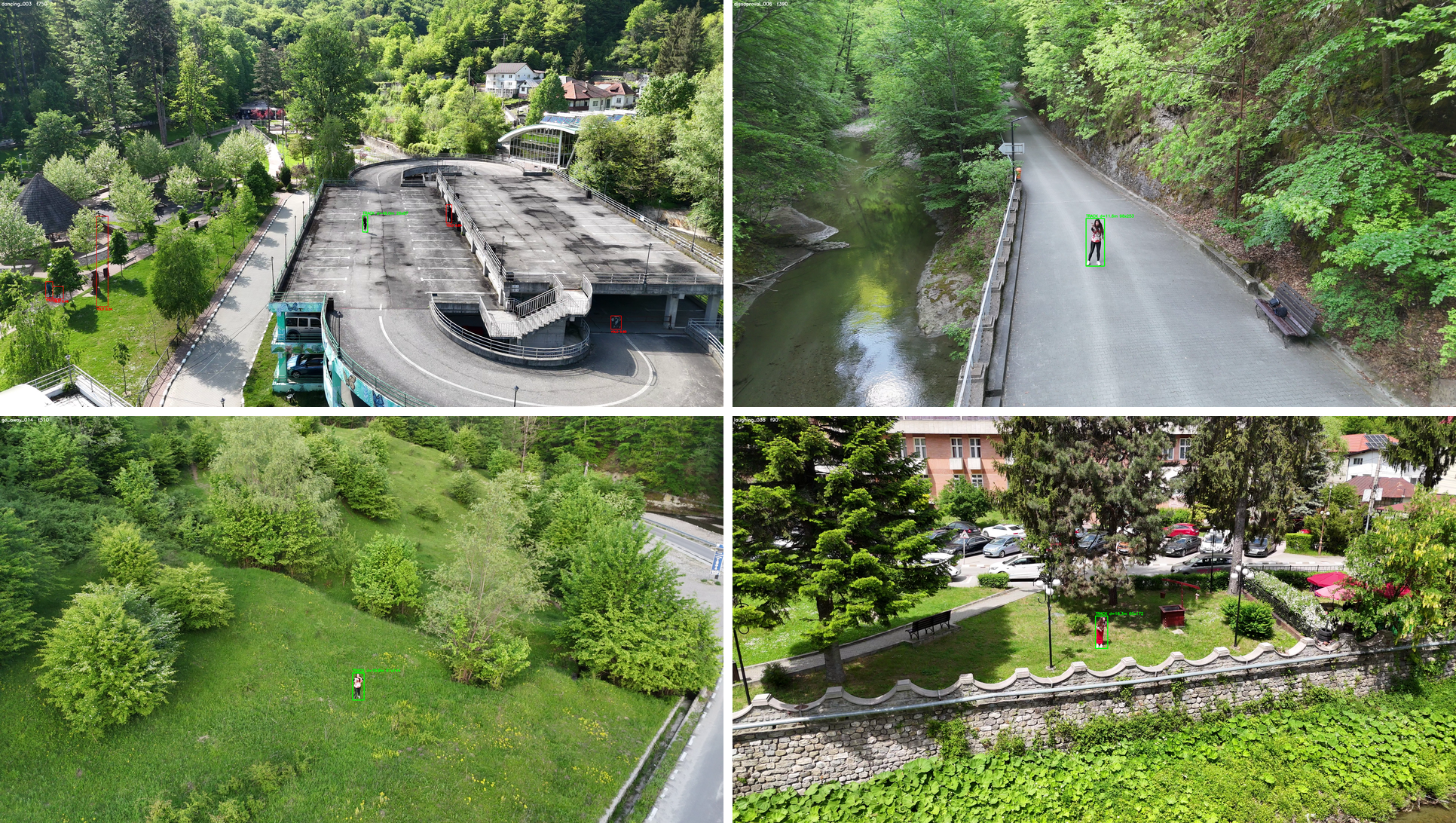}
    \vspace{1mm}
    \caption{Ground-truth (green) versus YOLO (red) detections on four aerial sequences from the \textit{Resort} scenes. The participant is instructed to show a single emotion, while the camera distance increases. Detection, keypoint visibility and classification performance can be achieved }
      \vspace{1mm}

    \label{fig:gt_vs_yolo_resort}
  \end{figure}

\section{Conclusions}

We set out to test whether ten intent-bearing body-language acts,
performed once by motion-captured actors and then transferred to
arbitrary characters using arbitrary drone footage can be reliably recovered and classified from raw aerial
footage and up to what distance. To support this, we will release a public real data benchmark and the tools needed to start from real or synthetic recordings and get to photorealistic placement in any scene, along with several drone case studies in
\textsc{Park} (\textit{placed} actors on a 4 minutes 4K clip) and
\textsc{Resort} (\textit{real} actor, 9 intents across 9 clips). Across
this benchmark we report (i)~per-distance YOLO11x-pose detection
rate, keypoint visibility and classifcation performance
(Fig.~\ref{fig:park-yolo-perf}, Fig.~\ref{fig:resort-yolo-perf}),
(ii)~classification accuracy together with selective-classification
risk-coverage curves on three large test sets
(Tab.~\ref{tab:consistency-top4}, Fig.~\ref{fig:risk-coverage}), and
(iii)~a label-free self-consistency score, $\cten$ usable as a
deployment-time trustworthiness proxy.

Three findings stand out:
\begin{itemize}
\setlength{\itemsep}{2pt}
\item \textbf{Pose-driven body-language communication is operationally
viable on real-world footage.} On placed actors a tiled detector
exceeds $0.9$ recall and recovers $\geq\!15$ of the $17$~COCO
keypoints out to ${\sim}50$\,m drone-to-subject distance, after which
both metrics drop sharply and collapse beyond $100$\,m; the
real-person \textsc{Resort} study confirms a track-and-classify
pipeline with $0.97$ average tracker retention inside the
${\sim}\!65$\,m operating envelope captured in these takes. Synthetic
and real placements degrade with depth, validating the
synthetic compositing as a controllable proxy for hard-to-collect
long-range real data.

\item \textbf{A label-free reliability signal can be measured on
deployment data alone.} Among the supervised reliability metrics
we evaluated, the sliding-window self-consistency score $\cten$ is one of the few that stays informative across all three test sets
($0.60$--$0.82$), complementing AURC, ECE and Brier with a measure
that needs no ground-truth labels. This lets a fielded system flag
its own degradation in the open-domain regime where calibration
metrics either saturate (MotionLCM) or become unreliable (VEO3.1).

\item \textbf{The benchmark opens immediate downstream research
directions.} The same pose recovery + emotion classification stack
underpins (a) covert ground-to-air signalling for emergency
communication, (b) search-and-rescue intent recognition where a
distressed subject can express need without colour-coded panels or
audio, and (c) entertainment / live-event analytics where crowd
gesture trends can be summarised without identifying individuals.
The distance envelope reported in this paper bounds where each of
these can be deployed today, and the public benchmark gives the
community a reproducible target for closing the long-range gap.
\end{itemize}

In short, body-language can be communicated from a single actor to
arbitrary observers, recovered from drone video with quantifiable
reliability inside a now-characterised distance envelope, and audited
in deployment without labels~--~a combination we believe is necessary
for any safety-critical use of pose-based behaviour understanding.

\noindent\textbf{Limitations.} Our study has clear boundaries. (1) It relies exclusively on 2D pose. We discard depth and, with it, motion toward and away from the camera; some intent confusions (\emph{come to me} vs.\ \emph{go away}) may be partly a consequence of this projection rather than of the models. (2) Our dataset is recorded in conditions favorable to keypoint detection, with a small number of participants. (3) The real-world study above is deliberately limited in subjects, scenes, and distance range, and genuine long-range drone footage processed through the full perception pipeline is limited, so deployment performance under real detector noise remains to be measured at scale.

\vspace{0.5em}

\noindent\textbf{Future work.} The natural next steps are 3D or multi-view pose to resolve depth-ambiguous intents, and a larger evaluation on real long-range captured footage to close the gap between benchmark and deployment. The label-free reliability signal also opens a practical avenue: using self-consistency as an on-device gate that abstains or defers when the input is not learnable, turning the analysis of this paper into a deployment-time safeguard.

\bibliography{egbib}

\begin{thebibliography}{50}
\providecommand{\natexlab}[1]{#1}
\providecommand{\url}[1]{\texttt{#1}}
\expandafter\ifx\csname urlstyle\endcsname\relax
  \providecommand{\doi}[1]{doi: #1}\else
  \providecommand{\doi}{doi: \begingroup \urlstyle{rm}\Url}\fi

\bibitem[Agrawal et~al.(2025)]{agrawal2025seamless}
Vasu Agrawal et~al.
\newblock Seamless interaction: Dyadic audiovisual motion modeling and large-scale dataset.
\newblock \emph{arXiv preprint arXiv:2506.22554}, 2025.

\bibitem[B{\"a}nziger et~al.(2012)B{\"a}nziger, Mortillaro, and Scherer]{banziger2012gemep}
Tanja B{\"a}nziger, Marcello Mortillaro, and Klaus~R Scherer.
\newblock Introducing the {Geneva Multimodal Expression} corpus for experimental research on emotion perception.
\newblock \emph{Emotion}, 12\penalty0 (5):\penalty0 1161--1179, 2012.

\bibitem[Bouazizi et~al.(2022)Bouazizi, Holzbock, Kressel, Dietmayer, and Belagiannis]{bouazizi2022motionmixer}
Arij Bouazizi, Adrian Holzbock, Ulrich Kressel, Klaus Dietmayer, and Vasileios Belagiannis.
\newblock {MotionMixer}: Mlp-based 3d human body pose forecasting.
\newblock In \emph{Proceedings of the International Joint Conference on Artificial Intelligence (IJCAI)}, 2022.

\bibitem[Busso et~al.(2008)Busso, Bulut, Lee, Kazemzadeh, Mower, Kim, Chang, Lee, and Narayanan]{busso2008iemocap}
Carlos Busso, Murtaza Bulut, Chi-Chun Lee, Abe Kazemzadeh, Emily Mower, Samuel Kim, Jeannette~N Chang, Sungbok Lee, and Shrikanth~S Narayanan.
\newblock Iemocap: Interactive emotional dyadic motion capture database.
\newblock \emph{Language resources and evaluation}, 42:\penalty0 335--359, 2008.

\bibitem[Cafaro et~al.(2016)Cafaro, Vilhj{\'a}lmsson, and Bickmore]{cafaro2016first}
Angelo Cafaro, Hannes~H{\"o}gni Vilhj{\'a}lmsson, and Timothy Bickmore.
\newblock First impressions in human--agent virtual encounters.
\newblock \emph{ACM Transactions on Computer-Human Interaction (TOCHI)}, 23\penalty0 (4):\penalty0 1--40, 2016.

\bibitem[Carion et~al.(2025)Carion, Gustafson, Hu, Debnath, Hu, Sur{\'i}s, Ryali, et~al.]{carion2025sam3}
Nicolas Carion, Laura Gustafson, Yuan-Ting Hu, Shoubhik Debnath, Ronghang Hu, Didac Sur{\'i}s, Chaitanya Ryali, et~al.
\newblock {SAM 3}: Segment anything with concepts.
\newblock Technical report, Meta Superintelligence Labs, November 2025.
\newblock URL \url{https://ai.meta.com/research/publications/sam-3-segment-anything-with-concepts/}.

\bibitem[Cauchard et~al.(2015)Cauchard, E, Zhai, and Landay]{cauchard2015drone}
Jessica~R Cauchard, Jane~L E, Kevin~Y Zhai, and James~A Landay.
\newblock Drone \& me: An exploration into natural human--drone interaction.
\newblock In \emph{Proceedings of the ACM International Joint Conference on Pervasive and Ubiquitous Computing (UbiComp)}, pages 361--365. ACM, 2015.

\bibitem[Chen et~al.(2021)Chen, Zhang, Yuan, Li, Deng, and Hu]{chen2021ctrgcn}
Yuxin Chen, Ziqi Zhang, Chunfeng Yuan, Bing Li, Ying Deng, and Weiming Hu.
\newblock Channel-wise topology refinement graph convolution for skeleton-based action recognition.
\newblock In \emph{ICCV}, pages 13359--13368, 2021.

\bibitem[Chi et~al.(2025)Chi, Chi, Chan, and Ramani]{chi2025infogcnpp}
Hyung-gun Chi, Seunggeun Chi, Stanley Chan, and Karthik Ramani.
\newblock {InfoGCN++}: Learning representation by predicting the future for online skeleton-based action recognition.
\newblock \emph{IEEE Transactions on Pattern Analysis and Machine Intelligence (TPAMI)}, 2025.

\bibitem[Cristescu and Isbister(2008)]{cristescu2008emotions}
Teodora Cristescu and Katherine Isbister.
\newblock Emotions and non-verbal behaviour in human-computer interactions.
\newblock In \emph{2008 10th International Conference on Development and Application Systems}, pages 68--73. IEEE, 2008.

\bibitem[Dai et~al.(2024)Dai, Chen, Wang, Liu, Dai, and Tang]{dai2024motionlcm}
Wenxun Dai, Ling-Hao Chen, Jingbo Wang, Jinpeng Liu, Bo~Dai, and Yansong Tang.
\newblock Motionlcm: Real-time controllable motion generation via latent consistency model.
\newblock \emph{arXiv preprint arXiv:2404.19759}, 2024.

\bibitem[Duan et~al.(2022)Duan, Wang, Chen, and Lin]{duan2022pyskl}
Haodong Duan, Jiaqi Wang, Kai Chen, and Dahua Lin.
\newblock {PYSKL}: Towards good practices for skeleton action recognition.
\newblock In \emph{Proceedings of the ACM International Conference on Multimedia (ACM MM)}, 2022.

\bibitem[Gal and Ghahramani(2016)]{gal2016dropout}
Yarin Gal and Zoubin Ghahramani.
\newblock Dropout as a {Bayesian} approximation: Representing model uncertainty in deep learning.
\newblock In \emph{ICML}, pages 1050--1059, 2016.

\bibitem[Geifman and El-Yaniv(2017)]{geifman2017selective}
Yonatan Geifman and Ran El-Yaniv.
\newblock Selective classification for deep neural networks.
\newblock In \emph{NeurIPS}, pages 4878--4887, 2017.

\bibitem[Geifman et~al.(2019)Geifman, Uziel, and El-Yaniv]{geifman2019bias}
Yonatan Geifman, Guy Uziel, and Ran El-Yaniv.
\newblock Bias-reduced uncertainty estimation for deep neural classifiers.
\newblock In \emph{International Conference on Learning Representations (ICLR)}, 2019.

\bibitem[{Google DeepMind}(2024)]{veo2024}
{Google DeepMind}.
\newblock Veo: a state-of-the-art model for video generation.
\newblock \url{https://deepmind.google/technologies/veo/}, May 2024.
\newblock Last accessed: 2026-02-12.

\bibitem[{Google DeepMind}(2025)]{deepmind2025veo}
{Google DeepMind}.
\newblock {Veo 3.1}: Improved text-to-video generation with native audio and narrative control.
\newblock \url{https://developers.googleblog.com/introducing-veo-3-1-and-new-creative-capabilities-in-the-gemini-api/}, October 2025.
\newblock Released 15 October 2025 via the Gemini API.

\bibitem[Guo et~al.(2017)Guo, Pleiss, Sun, and Weinberger]{guo2017calibration}
Chuan Guo, Geoff Pleiss, Yu~Sun, and Kilian~Q Weinberger.
\newblock On calibration of modern neural networks.
\newblock In \emph{ICML}, pages 1321--1330, 2017.

\bibitem[Hazmoune(2024)]{hazmoune2024transformers}
Soufiane Hazmoune.
\newblock Using transformers for multimodal emotion recognition.
\newblock \emph{Engineering Applications of Artificial Intelligence}, 133, 2024.

\bibitem[Jocher and Qiu(2024)]{jocher2024yolo11}
Glenn Jocher and Jing Qiu.
\newblock {Ultralytics YOLO11}, 2024.
\newblock URL \url{https://github.com/ultralytics/ultralytics}.

\bibitem[Karaev et~al.(2024)Karaev, Rocco, Graham, Neverova, Vedaldi, and Rupprecht]{karaev2024cotracker}
Nikita Karaev, Ignacio Rocco, Benjamin Graham, Natalia Neverova, Andrea Vedaldi, and Christian Rupprecht.
\newblock Cotracker: It is better to track together.
\newblock In \emph{European conference on computer vision}, pages 18--35. Springer, 2024.

\bibitem[Kim et~al.(2025)]{kim2025drone}
Jeonghyeon Kim et~al.
\newblock A human-following drone providing gesture recognition to control {IoT} devices based on {3D} body-landmark detection.
\newblock \emph{International Journal of Control, Automation and Systems}, 2025.
\newblock \doi{10.1007/s12555-025-0012-y}.

\bibitem[Lin et~al.(2014)Lin, Maire, Belongie, Hays, Perona, Ramanan, Doll{\'a}r, and Zitnick]{lin2014microsoft}
Tsung-Yi Lin, Michael Maire, Serge Belongie, James Hays, Pietro Perona, Deva Ramanan, Piotr Doll{\'a}r, and C~Lawrence Zitnick.
\newblock Microsoft coco: Common objects in context.
\newblock In \emph{European conference on computer vision}, pages 740--755. Springer, 2014.

\bibitem[Liu et~al.(2020)Liu, Shahroudy, Perez, Wang, Duan, and Kot]{liu2020ntu120}
Jun Liu, Amir Shahroudy, Mauricio Perez, Gang Wang, Ling-Yu Duan, and Alex~C Kot.
\newblock {NTU RGB+D} 120: A large-scale benchmark for {3D} human activity understanding.
\newblock \emph{IEEE Transactions on Pattern Analysis and Machine Intelligence}, 42\penalty0 (10):\penalty0 2684--2701, 2020.

\bibitem[Locke(2011)]{locke2011circumplex}
Kenneth~D. Locke.
\newblock Circumplex measures of interpersonal constructs.
\newblock In Leonard~M. Horowitz and Stephen Strack, editors, \emph{Handbook of Interpersonal Psychology: Theory, Research, Assessment, and Therapeutic Interventions}, pages 313--324. Wiley, Hoboken, NJ, 2011.
\newblock \doi{10.1002/9781118001868.ch19}.

\bibitem[Lozano et~al.(2023)Lozano, S{\'a}nchez-Torres, L{\'o}pez-Nava, and Favela]{lozano2023open}
Ernesto~A Lozano, Carlos~E S{\'a}nchez-Torres, Irvin~H L{\'o}pez-Nava, and Jes{\'u}s Favela.
\newblock An open framework for nonverbal communication in human-robot interaction.
\newblock In \emph{International Conference on Ubiquitous Computing and Ambient Intelligence}, pages 21--32. Springer, 2023.

\bibitem[Ma et~al.(2022)Ma, Nie, Long, Zhang, and Li]{ma2022pgbig}
Tiezheng Ma, Yongwei Nie, Chengjiang Long, Qing Zhang, and Guiqing Li.
\newblock Progressively generating better initial guesses towards next stages for high-quality human motion prediction.
\newblock In \emph{Proceedings of the IEEE/CVF Conference on Computer Vision and Pattern Recognition (CVPR)}, 2022.

\bibitem[Mahmood et~al.(2019)Mahmood, Ghorbani, Farsa, Tuzel, and Black]{mahmoud2019amass}
Naureen Mahmood, Nima Ghorbani, Nikolaus Farsa, Oncel Tuzel, and Michael~J Black.
\newblock Amass: Archive of motion capture as surface shape.
\newblock In \emph{Proceedings of the IEEE/CVF International Conference on Computer Vision}, pages 5442--5451, 2019.

\bibitem[Mart{\'i}nez-Gonz{\'a}lez et~al.(2021)Mart{\'i}nez-Gonz{\'a}lez, Villamizar, and Odobez]{martinez2021potr}
Angel Mart{\'i}nez-Gonz{\'a}lez, Michael Villamizar, and Jean-Marc Odobez.
\newblock Pose transformers ({POTR}): Human motion prediction with non-autoregressive transformers.
\newblock In \emph{Proceedings of the IEEE/CVF International Conference on Computer Vision Workshops (ICCVW)}, 2021.

\bibitem[Noroozi et~al.(2021)Noroozi, Kaminska, Corneanu, Sapinski, Escalera, and Anbarjafari]{noroozi2021emotional}
Fatemeh Noroozi, Dorota Kaminska, Ciprian Corneanu, Tomasz Sapinski, Sergio Escalera, and Gholamreza Anbarjafari.
\newblock Survey on emotional body gesture recognition.
\newblock \emph{IEEE Transactions on Affective Computing}, 12\penalty0 (2):\penalty0 505--523, 2021.

\bibitem[Paszke et~al.(2017)Paszke, Gross, Chintala, Chanan, Yang, DeVito, Lin, Desmaison, Antiga, and Lerer]{paszke2017automatic}
Adam Paszke, Sam Gross, Soumith Chintala, Gregory Chanan, Edward Yang, Zachary DeVito, Zeming Lin, Alban Desmaison, Luca Antiga, and Adam Lerer.
\newblock Automatic differentiation in pytorch, 2017.

\bibitem[Pavlakos et~al.(2019)Pavlakos, Choutas, Ghorbani, Bolkart, Osman, Tzionas, and Black]{pavlakos2019smplx}
Georgios Pavlakos, Vasileios Choutas, Nima Ghorbani, Timo Bolkart, Ahmed A.~A. Osman, Dimitrios Tzionas, and Michael~J. Black.
\newblock Expressive body capture: {3D} hands, face, and body from a single image.
\newblock In \emph{IEEE/CVF Conference on Computer Vision and Pattern Recognition (CVPR)}, pages 10975--10985, 2019.

\bibitem[Perera et~al.(2021)Perera, Law, Ogunwa, and Chahl]{perera2021uav}
Asanka~G Perera, Yee~Wei Law, Titilayo~T Ogunwa, and Javaan Chahl.
\newblock Real-time human detection and gesture recognition for on-board {UAV} rescue.
\newblock \emph{Sensors}, 21\penalty0 (6):\penalty0 2180, 2021.

\bibitem[Poria et~al.(2018)Poria, Hazarika, Majumder, Naik, Cambria, and Mihalcea]{poria2018meld}
Soujanya Poria, Devamanyu Hazarika, Navonil Majumder, Gautam Naik, Erik Cambria, and Rada Mihalcea.
\newblock Meld: A multimodal multi-party dataset for emotion recognition in conversations.
\newblock \emph{arXiv preprint arXiv:1810.02508}, 2018.

\bibitem[Rempe et~al.(2026)Rempe, Petrovich, Yuan, Zhang, Peng, Jiang, Wang, Iqbal, Minor, de~Ruyter, Li, Tessler, Lim, Jeong, Wu, Hassani, Huang, Yu, Chung, Song, Dionne, Kautz, Yuen, and Fidler]{rempe2026kimodo}
Davis Rempe, Mathis Petrovich, Ye~Yuan, Haotian Zhang, Xue~Bin Peng, Yifeng Jiang, Tingwu Wang, Umar Iqbal, David Minor, Michael de~Ruyter, Jiefeng Li, Chen Tessler, Edy Lim, Eugene Jeong, Sam Wu, Ehsan Hassani, Michael Huang, Jin-Bey Yu, Chaeyeon Chung, Lina Song, Olivier Dionne, Jan Kautz, Simon Yuen, and Sanja Fidler.
\newblock {Kimodo}: Scaling controllable human motion generation.
\newblock \emph{arXiv preprint arXiv:2603.15546}, 2026.
\newblock NVIDIA Toronto AI Lab technical report. Project page: \url{https://research.nvidia.com/labs/sil/projects/kimodo/}.

\bibitem[Rivera et~al.(2024)Rivera, Rodrigues, and Fugita]{rivera2024emotion}
Fernando~Pujaico Rivera, Paulo~Sergio Rodrigues, and Oscar Eduardo~Hidetoshi Fugita.
\newblock Emotion recognition from facial images, body gestures, and skeletal posture keypoints: The ber2024 dataset.
\newblock \emph{Computer Methods and Programs in Biomedicine}, 2024.

\bibitem[Saunderson et~al.(2019)Saunderson, Nehaniv, and Dautenhahn]{saunderson2019robots}
Sam Saunderson, Chrystopher~L Nehaniv, and Kerstin Dautenhahn.
\newblock Robots with different non-verbal communication styles: How the way a robot gestures, moves and looks affects people’s ratings of the robot’s social qualities.
\newblock In \emph{Companion of the 2019 ACM/IEEE International Conference on Human-Robot Interaction}, pages 551--559, 2019.

\bibitem[Shi et~al.(2019{\natexlab{a}})Shi, Zhang, Cheng, and Lu]{shi2019twostream}
Lei Shi, Yifan Zhang, Jian Cheng, and Hanqing Lu.
\newblock Two-stream adaptive graph convolutional networks for skeleton-based action recognition.
\newblock In \emph{Proceedings of the IEEE/CVF Conference on Computer Vision and Pattern Recognition (CVPR)}, pages 12026--12035, 2019{\natexlab{a}}.

\bibitem[Shi et~al.(2019{\natexlab{b}})Shi, Zhang, Cheng, and Lu]{shi2019twostreamadaptive}
Lei Shi, Yifan Zhang, Jian Cheng, and Hanqing Lu.
\newblock Two-stream adaptive graph convolutional networks for skeleton-based action recognition.
\newblock In \emph{CVPR}, pages 12026--12035, 2019{\natexlab{b}}.

\bibitem[Shin et~al.(2025)Shin, Evetts, Saylor, Kim, Woo, Rhee, and Kim]{skybound2025}
Soohyun Shin, Trevor Evetts, Hunter Saylor, Hyunji Kim, Soojin Woo, Wonhwha Rhee, and Seong-Woo Kim.
\newblock Skybound magic: Enabling body-only drone piloting through a lightweight vision--pose interaction framework.
\newblock \emph{International Journal of Human--Computer Interaction}, 2025.
\newblock \doi{10.1080/10447318.2025.2546039}.

\bibitem[Song et~al.(2022)Song, Zhang, Shan, and Wang]{song2022efficientgcn}
Yi-Fan Song, Zhang Zhang, Caifeng Shan, and Liang Wang.
\newblock Constructing stronger and faster baselines for skeleton-based action recognition.
\newblock \emph{IEEE Transactions on Pattern Analysis and Machine Intelligence (TPAMI)}, 2022.

\bibitem[Suarez-Fernandez et~al.(2016)Suarez-Fernandez, Sanchez-Lopez, Sampedro, Bavle, Molina, and Campoy]{suarez2016natural}
Roberto Suarez-Fernandez, Jose~Luis Sanchez-Lopez, Carlos Sampedro, Hriday Bavle, Martin Molina, and Pascual Campoy.
\newblock Natural user interfaces for human-drone multi-modal interaction.
\newblock In \emph{International Conference on Unmanned Aircraft Systems (ICUAS)}, pages 1013--1022. IEEE, 2016.

\bibitem[Urakami and Seaborn(2023)]{urakami2023nonverbal}
Jacqueline Urakami and Katie Seaborn.
\newblock Nonverbal cues in human--robot interaction: A communication studies perspective.
\newblock \emph{ACM Transactions on Human-Robot Interaction}, 12\penalty0 (2):\penalty0 1--21, 2023.

\bibitem[Wang et~al.(2025)Wang, Xu, Dong, Deng, Xiang, Lv, Sun, Tong, and Yang]{wang2025moge2}
Ruicheng Wang, Sicheng Xu, Yue Dong, Yu~Deng, Jianfeng Xiang, Zelong Lv, Guangzhong Sun, Xin Tong, and Jiaolong Yang.
\newblock {MoGe-2}: Accurate monocular geometry with metric scale and sharp details.
\newblock In \emph{Advances in Neural Information Processing Systems (NeurIPS)}, 2025.

\bibitem[Wiggins(1979)]{wiggins1979psychology}
Jerry~S. Wiggins.
\newblock A psychological taxonomy of trait-descriptive terms: The interpersonal domain.
\newblock \emph{Journal of Personality and Social Psychology}, 37\penalty0 (3):\penalty0 395--412, 1979.
\newblock \doi{10.1037/0022-3514.37.3.395}.

\bibitem[Xu et~al.(2024)Xu, Zhang, Zhang, and Tao]{xu2024vitpose}
Yufei Xu, Jing Zhang, Qiming Zhang, and Dacheng Tao.
\newblock {ViTPose++}: Vision transformer for generic body pose estimation.
\newblock \emph{IEEE Transactions on Pattern Analysis and Machine Intelligence}, 46\penalty0 (2):\penalty0 1212--1230, 2024.

\bibitem[Yan et~al.(2018)Yan, Xiong, and Lin]{yan2018stgcn}
Sijie Yan, Yuanjun Xiong, and Dahua Lin.
\newblock Spatial temporal graph convolutional networks for skeleton-based action recognition.
\newblock In \emph{AAAI}, pages 7444--7452, 2018.

\bibitem[Zadeh et~al.(2018)Zadeh, Liang, Mazumder, Cambria, and Morency]{zadeh2018mosei}
Amir Zadeh, Paul~Pu Liang, Soujanya~Poria Mazumder, Erik Cambria, and Louis-Philippe Morency.
\newblock Multimodal language analysis in the wild: {CMU-MOSEI} dataset and interpretable dynamic fusion graph.
\newblock In \emph{ACL}, pages 2236--2246, 2018.

\bibitem[Zanfir et~al.(2013)Zanfir, Leordeanu, and Sminchisescu]{zanfir2013moving}
Mihai Zanfir, Marius Leordeanu, and Cristian Sminchisescu.
\newblock The moving pose: An efficient 3d kinematics descriptor for low-latency action recognition and detection.
\newblock In \emph{Proceedings of the IEEE international conference on computer vision}, pages 2752--2759, 2013.

\bibitem[Zhu et~al.(2023)Zhu, Ma, Liu, Liu, Wu, and Wang]{zhu2023motionbert}
Wentao Zhu, Xiaoxuan Ma, Zhaoyang Liu, Libin Liu, Wayne Wu, and Yizhou Wang.
\newblock {MotionBERT}: A unified perspective on learning human motion representations.
\newblock In \emph{ICCV}, pages 15085--15099, 2023.

\end{thebibliography}

\end{document}